\documentclass[journal]{IEEEtran}
\usepackage{cite}
\usepackage{graphics}
\usepackage{graphicx}
\usepackage{amsmath,amsthm,amssymb}
\usepackage{algorithm}
\usepackage{algorithmic}
\usepackage{caption}
\usepackage{amsfonts}
\usepackage{multirow}
\usepackage[dvipsnames]{xcolor}
\usepackage{subfigure}
\usepackage{booktabs}
\usepackage{palatino}
\usepackage{tikz}
\usepackage{latexsym}
\usetikzlibrary{shapes.geometric, arrows}
\renewcommand\arraystretch{1.8}

\begin{document}

\title{Unsupervised Green Object Tracker (GOT) without Offline Pre-training}

\author{Zhiruo~Zhou,~\IEEEmembership{Student~member,~IEEE,} Suya~You,
        and~C.-C.~Jay~Kuo,~\IEEEmembership{Fellow,~IEEE}
\thanks{Zhiruo Zhou and C.-C. Jay Kuo are with the Ming-Hsieh
Department of Electrical and Computer Engineering, University of
Southern California, CA, 90089-2564, USA (e-mails: zhiruozh@usc.edu and cckuo@sipi.usc.edu).}%
\thanks{Suya You is with Army Research Laboratory, Adelphi,
Maryland, USA (e-mail: suya.you.civ@army.mil).}%
\thanks{Distribution Statement A: Approved for public release. Distribution is unlimited.}
}

\maketitle

\begin{abstract}

Supervised trackers trained on labeled data dominate the single object
tracking field for superior tracking accuracy. The labeling cost and the
huge computational complexity hinder their applications on edge devices.
Unsupervised learning methods have also been investigated to reduce the
labeling cost but their complexity remains high. Aiming at lightweight
high-performance tracking, feasibility without offline pre-training, and
algorithmic transparency, we propose a new single object tracking
method, called the green object tracker (GOT), in this work. GOT
conducts an ensemble of three prediction branches for robust box
tracking: 1) a global object-based correlator to predict the object
location roughly, 2) a local patch-based correlator to build temporal
correlations of small spatial units, and 3) a superpixel-based
segmentator to exploit the spatial information of the target frame.  GOT
offers competitive tracking accuracy with state-of-the-art unsupervised
trackers, which demand heavy offline pre-training, at a lower
computation cost. GOT has a tiny model size ($<$3k parameters) and low
inference complexity (around 58M FLOPs per frame).  Since its inference
complexity is between $0.1\%\sim10\%$ of DL trackers, it can be easily
deployed on mobile and edge devices. 

\end{abstract}


\section{Introduction}\label{sec:introduction}

\IEEEPARstart{V}{ideo} object tracking is one of the fundamental
computer vision problems \cite{javed2022visual} and finds applications
in various applications such as autonomous driving
\cite{lee2015road,janai2020computer} and video surveillance
\cite{xing2010multiple}. Given the ground-truth bounding box of the
object in the first frame of a test video, a single object tracker (SOT)
predicts the object box in all subsequent frames.  Most trackers follow
the tracking-by-detection paradigm. That is, based on the object
template obtained in the $(i-1)$th frame (i.e., the reference frame), a
tracker conducts similarity matching over a search region at the $i$th
frame (i.e., the target frame).  This setting is used to reflect an
online real-time tracking environment, where the data processing is
applied to streaming video with a small memory buffer. 

Research on SOT has a long history, which will be briefly reviewed in
Sec.  \ref{sec:review}. There are two major breakthroughs in SOT
development. The first one lies in the use of the discriminative
correlation filter (DCF) \cite{bolme2010visual} and its variants. Based
on handcrafted features (e.g., the histogram of oriented gradients (HOG)
and colornames (CN) \cite{danelljan2015learning}) extracted from the
reference template, DCF trackers estimate the location and size of the
target template by examining the correlation (or similarity) between the
reference template and the image content in the target search region.
The second one arises by exploiting deep neural networks (DNNs) or deep
learning (DL).  Supervised and unsupervised DL trackers with pre-trained
networks have been dominating in their respective categories in recent
years.  They are trained with large-scale offline pre-training data. The
former has human-labeled object boxes throughout all frames, while the
latter does not, in all training sequences. 

There is a link between DCF and DL trackers.  One representative branch
of supervised DL trackers is known as the Siamese network, which
maintains the template matching idea. On the other hand, DL trackers
adopt the end-to-end optimization approach to derive powerful deep
features for the matching purpose. Besides the backbone network, they
incorporate several auxiliary subnetworks called heads, e.g., the
classification head and the box regression head. 

The superior tracking accuracy of supervised DL trackers is attributed
to a huge amount of efforts in offline pre-training with densely labeled
videos and images. In addition, the backbone network gets larger and
larger from the AlexNet to the Transformer. Generally speaking, DL
trackers demand a large model size and high computational complexity.
The heavy computational burden hinders their practical applications in
edge devices. For example, SiamRPN++ \cite{li2019siamrpn++} has a model
containing 54M parameters and takes 48.9G floating point operations
(flops) to track one frame. To lower the high computational resource
requirement, research has been done to compress the model via neural
architecture search \cite{yan2021lighttrack}, model distillation
\cite{shen2021distilled}, or networks pruning and quantization
\cite{blatter2023efficient,chen2022efficient, borsuk2022fear,
jung2022online,aggarwal2023designing}. 

\begin{figure}[t]
\centerline{\includegraphics[width=\linewidth]{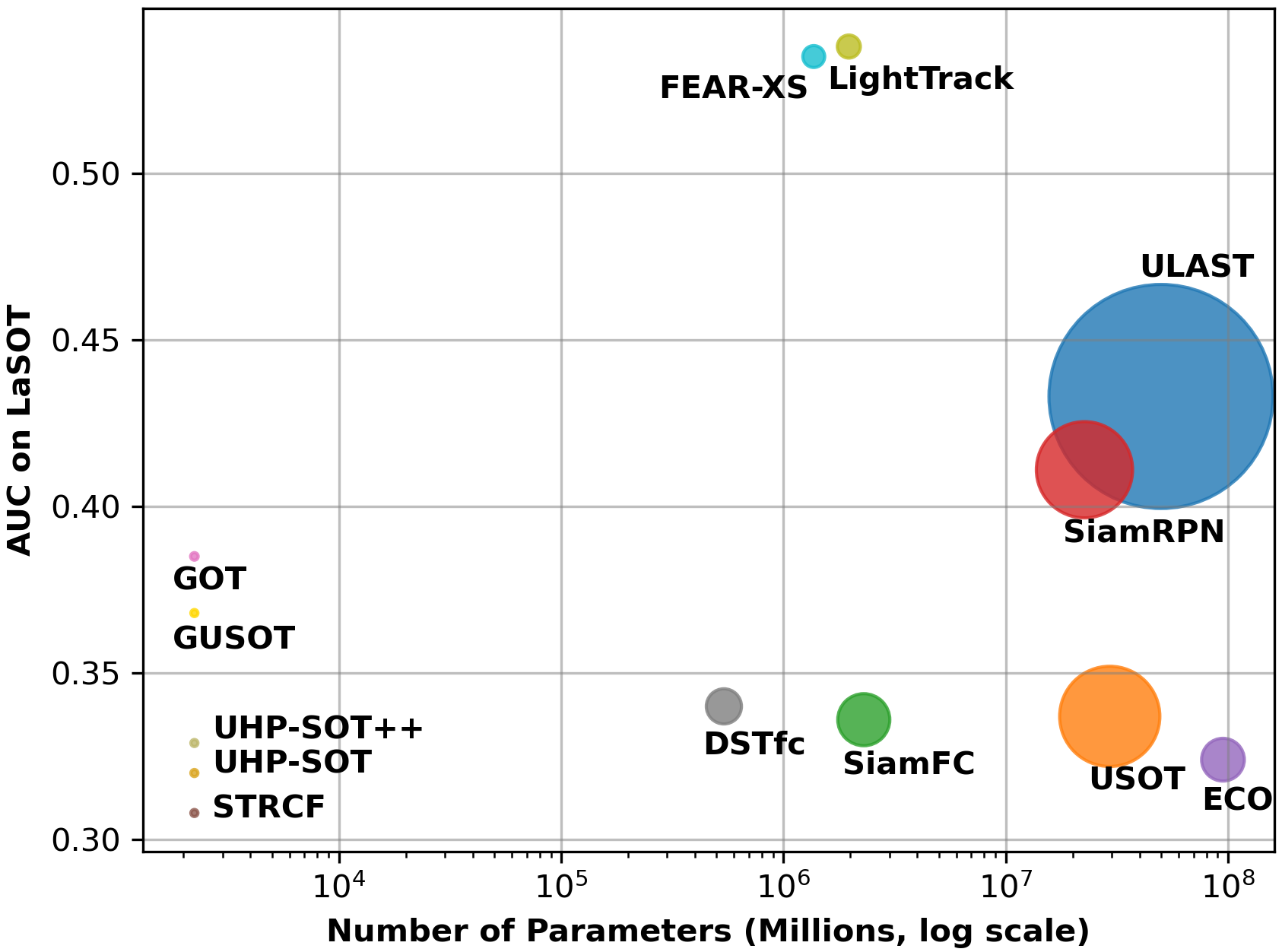}}
\caption{Comparison of object trackers in the number of model parameters
(along the x-axis), the AUC performance (along the y-axis) and inference
complexity in floating point operations (in circle sizes) with respect to 
the LaSOT dataset.} \label{fig:bubble}
\end{figure}

One recent research activity lies in reducing the labeling cost.
Along this line, unsupervised DL trackers have been proposed to enable
intelligent learning, e.g., \cite{wang2021unsupervisedcvpr,
wu2021progressive, zheng2021learning, shen2022unsupervised}. In the
training process, they generate pseudo object boxes in initial frames, allow a tracker to track in both forward and backward directions,
and enforce the cycle consistency.  Various techniques have been
proposed to adjust pseudo labels and improve training efficiency.
Unsupervised DL trackers contain complicated networks needed for large-scale offline pre-training, leading to large model sizes. The
state-of-the-art unsupervised DL tracker, ULAST
\cite{shen2022unsupervised}, achieves comparable performance as top
supervised DL trackers. As a modification of SiamRPN++, ULAST has a
large model size and heavy computational complexity in inference. 

Our research goal is to develop unsupervised, high-performance, and
lightweight trackers, where lightweightness is measured in model sizes
and inference computational complexity. Toward this objective, we have
developed new trackers by extending DCF trackers. Examples include
UHP-SOT \cite{zhou2021uhp}, UHP-SOT++ \cite{zhou2022uhp} and GUSOT
\cite{zhou2022gusot}. The extensions include an object recovery
mechanism and flexible shape estimation in the face of occlusion and
deformation, respectively. They improved the tracking accuracy of DCF
trackers greatly while maintaining their lightweight advantage.  These
trackers were not only unsupervised but also demanded no offline
pre-training. Furthermore, these trackers adopted a modular design for
algorithmic transparency. 

Based on the above discussion, we can categorize object trackers into
three types according to their training strategies: A) supervised
trackers, B) unsupervised trackers with offline pre-training, and C)
unsupervised trackers without offline pre-training. In terms of training
complexity, Type B has the highest training complexity while Type C has
the lowest training complexity (which is almost none). We consider their
representative trackers in Fig. \ref{fig:bubble}:
\begin{itemize}
\item Type A: SiamFC, ECO, SiamRPN, LightTrack, DSTfc, and FEAR-XS; 
\item Type B: USOT and ULAST;
\item Type C: STRCF, UHP-SOT, UHP-SOT++, GUSOT, and GOT.
\end{itemize}
We compare their characteristics in three aspects in the figure:
tracking performance (along the y-axis), model sizes (along the x-axis),
and inference complexity (in circle sizes).  

The green object tracker (GOT) is a new tracker proposed in this work.
It is called ``green" due to its low computational complexity in both
training and inference stages, leading to a low carbon footprint. There
is an emerging research trend in artificial intelligence (AI) and
machine learning (ML) by taking the carbon footprint into account. It is
called ``green learning" \cite{kuo2022green}. Besides sustainability,
green learning emphasizes algorithmic transparency by adopting a modular
design. GOT has been developed based on the green learning principle.

GOT conducts an ensemble of three prediction branches for robust object
tracking: 1) a global object-based correlator to predict the object
location roughly, 2) a local patch-based correlator to build temporal
correlations of small spatial units, and 3) a superpixel-based
segmentator to exploit the spatial information (e.g., color similarity
and geometrical constraints) of the target frame.  For the first and the
main branch, GOT adopts GUSOT as the baseline. The outputs from three
branches are then fused to generate the ultimate object box, where an
innovative fusion strategy is developed. 

GOT contains two novel ideas that have been neglected in the existing 
object tracking literature. They are elaborated below. 
\begin{itemize}
\item The performance of the global correlator in the first branch
degrades when the tracked object has severe deformation between two
adjacent frames.  The local patch-based correlator in the second branch
is used to provide more flexible shape estimation and object
re-identification. It is essential to implement the local correlator
efficiently. It is formulated as a binary classification problem. It
classifies a local patch into one of two classes - belonging to the object
or the background. 

\item The tracking process usually alternates between the easy steady period and the challenging period as confronted with deformations and occlusion. The proposed fuser monitors the tracking quality and fuses different box proposals according to
the tracking dynamics to ensure robustness against challenges while maintaining a reasonable complexity.
\end{itemize}

We evaluate GOT on five benchmarking datasets for thorough performance
comparison. They are OTB2015, VOT2016, TrackingNet, LaSOT, and OxUvA.  It
is demonstrated by extensive experiments and ablation studies that GOT
offers competitive tracking accuracy with state-of-the-art unsupervised
trackers (i.e., USOT and ULAST), which demand heavy offline
pre-training, at a lower computation cost.  GOT has a tiny model size ($<$3k parameters) and low inference complexity (around 58M FLOPs per
frame). Its inference complexity is between $0.1\%\sim10\%$ of DL
trackers. Thus, it can be easily deployed on mobile and edge devices.
Furthermore, we discuss the role played by supervision and offline
pre-training to shed light on our design. 

The rest of this paper is organized as follows. Related work is reviewed
in Sec. \ref{sec:review}. The GOT method is proposed in Sec.
\ref{sec:method}. Experimental results are shown in Sec.
\ref{sec:experiments}. Concluding remarks are given in Sec.
\ref{sec:conclusion}. 

\section{Related Work}\label{sec:review}

\subsection{DCF Trackers}

Unsupervised DCF trackers without offline pre-training had been popular
before the arrival of DL trackers. Given a reference template, DCF
trackers conduct circulant patch sampling on the target frame and
predict the location and size of the object template via regression.
Quite a few DCF trackers with various regression objective functions or
feature representations were proposed, e.g., \cite{bolme2010visual,
henriques2014high, danelljan2015convolutional,
danelljan2016discriminative, danelljan2016beyond, bertinetto2016staple,
li2018learning, xu2019learning, li2020autotrack}. Classic DCF trackers
estimate the scale change by checking multiple scales.  Yet, they are
not flexible in adjusting the aspect ratio of the bounding box.
Recently, DCF-based trackers such as UHP-SOT++ \cite{zhou2022uhp} and
GUSOT \cite{zhou2022gusot} allow more flexible shape change by adopting
low-cost segmentation techniques and exploiting motion residuals.  The
latter can facilitate object re-identification after tracking loss.
Generally speaking, all DCF trackers meet the requirement of being an
unsupervised lightweight solution without offline pre-training. The main
concern is their poorer tracking accuracy as compared with modern DL
trackers. Thus, the main task is how to boost the tracking performance
with little extra cost in memory and computation. In this work, we adopt
GUSOT \cite{zhou2022gusot} as the baseline of GOT since it has
demonstrated good performance in tracking long videos.

\subsection{DL Trackers}

\subsubsection{Offline Pre-training}

The majority of high-performance DL trackers adopt offline pre-training
on large-scale datasets, including still images
\cite{deng2009imagenet,lin2014microsoft} and densely annotated videos
\cite{russakovsky2015imagenet,real2017youtube,muller2018trackingnet,
huang2019got}.  Modularized trackers use pre-trained convolutional
neural networks (CNNs) as the feature extraction backbone
\cite{ma2015hierarchical,qi2016hedged,danelljan2017eco}.  End-to-end
trackers need to finetune the backbone with auxiliary networks to be
adapted to the tracking task \cite{bertinetto2016fully,
li2018high,li2019siamrpn++, tao2016siamese, zhu2018distractor,
zhang2019deeper}. The transformer boosts the tracking accuracy of
supervised DL trackers to a higher level \cite{wang2021transformer,
chen2021transformer}. Although the power of offline pre-training with
annotated boxes in tracking performance boosting is obvious, there are
associated costs.  First, it is a heavy burden to scale up the training
data.  Second, one needs to remove noise in newly sourced videos to
obtain high-quality annotations. Third, the costly offline training
process yields a large carbon footprint. 

\subsubsection{Unsupervised Trackers}

To address the high human labeling cost, researchers investigate ways to
conduct offline pre-training with unlabeled data
\cite{wang2021unsupervisedcvpr, wu2021progressive, zheng2021learning,
shen2022unsupervised}. As proposed in ResPUL \cite{wu2021progressive},
one idea is to train the backbone network offline with contrastive
learning on static images and enhance the learning process with temporal
sample mining. Another idea is to impose cycle consistency in offline
pre-training. For example, UDT \cite{wang2021unsupervisedcvpr} proposed
the cycle training method. It randomly crops patches in the first frame
as object templates (or pseudo labels) and trains the tracker to track
forward and then backward to yield a consistent object location in the
initial frame. Later work put efforts into cleaning noisy pseudo labels
and improving cycle training efficiency. Rather than random cropping on
any video frame, USOT \cite{zheng2021learning} detected moving objects
using a dense optical flow and selected valid video segments to avoid
influence from occlusion or out-of-view. It also expanded the cycle
training interval.  As the state-of-the-art unsupervised tracker, ULAST
\cite{shen2022unsupervised} applied a region mask to
filter out possible contaminations from the non-object region and weigh the loss from pseudo labels of different quality.  ULAST
can achieve comparable performance against supervised trackers with a
large network and high computational complexity.  It demands large-scale
offline pre-training and its training complexity is significantly higher
than that of supervised trackers due to the extra cycle consistency
requirement. 

\subsubsection{Lightweight Trackers}

The majority of DL trackers rely on powerful yet heavy backbones (e.g.,
CNNs or transformers). To deploy them on resource-limited platforms,
efforts have been made in compressing a network without degrading its
tracking accuracy. Various approaches have been proposed such as neural
architecture search (NAS) \cite{yan2021lighttrack}, model distillation
\cite{shen2021distilled}, network quantization \cite{jung2022online},
feature sparsification, channel pruning \cite{aggarwal2023designing}, or
other specific designs to reduce the complexity of original network
layers \cite{chen2022efficient,blatter2023efficient,borsuk2022fear}. As
a pioneering method, LightTrack \cite{yan2021lighttrack} lays the
foundation for later lightweight trackers. LightTrack adopted NAS to
compress a large-size supervised tracker into its mobile counterpart.
Its design process involved the following three steps: 1) train a
supernet, 2) search for its optimal subnet, and 3) re-train and tune the
subnet on a large number of training data. The inference complexity can
be reduced to around 600M flops per frame at the end.  Simply speaking,
it begins with a well-designed supervised tracker and attempts to reduce
the model size and complexity with re-training.  Another example was
proposed in \cite{shen2021distilled}. It also conducted NAS to find a
small network and used it to distill the knowledge of a large-size
tracker via teacher-student training. 

Our work is completely different from DL trackers as the proposed GOT
does not have an end-to-end optimized neural network architecture. It
adopts a modularized and interpretable system design. It is unsupervised
without offline pre-training. It is proper to view GOT as a descent (or
a modern version) of classic DCF trackers.  Our main task in developing
GOT is to identify the shortcomings of classic DCF trackers and find
their remedies. 
\begin{figure*}[htbp]
\centerline{\includegraphics[width=\textwidth]{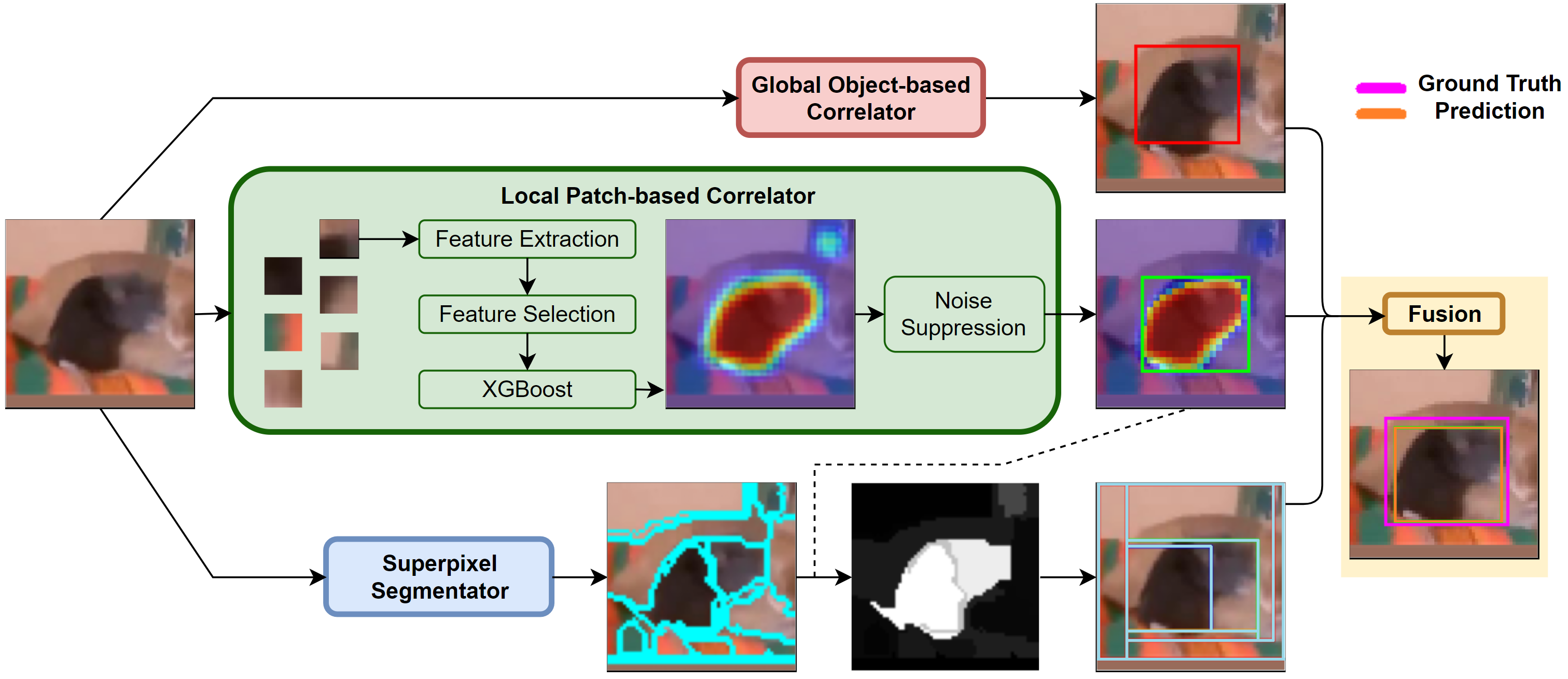}}
\caption{The system diagram of the proposed green object tracker 
(GOT). The global object-based correlator generates a rigid proposal,
while the local patch-based correlator outputs a deformable box and a
objectness score map which helps the segmentator calculate additional
deformable boxes. These proposals are fused into one final
prediction.}\label{fig:system}
\end{figure*}

\section{Green Object Tracker (GOT)}\label{sec:method}

The system diagram of the proposed green object tracker (GOT) is
depicted in Fig. \ref{fig:system}. It contains three bounding box
prediction branches: 1) a global object-based correlator, 2) a local
patch-based correlator applied to small spatial units, and 3) a
superpixel segmentator.  Each branch will offer one or multiple
proposals from the input search region, and they will be fused to yield
the final prediction. We use GUSOT \cite{zhou2022gusot} in the first
branch and the superpixel segmentation technique
\cite{felzenszwalb2004efficient} in the third branch. In the following,
we provide a brief review of the first branch in Sec. \ref{subsec:1st},
and elaborate on the second branch in Sec.  \ref{subsec:2nd}. We do not
spend any space on the superpixel segmentator since it is directly taken
from \cite{felzenszwalb2004efficient}.  Finally, we present the fusion
strategy in Sec. \ref{subsec:integration}. 

\subsection{Global Object-based Correlator}\label{subsec:1st}

The GUSOT tracker is the evolved result of a series of efforts in
enhancing the performance of lightweight DCF-based trackers. They
include STRCF \cite{li2018learning}, UHP-SOT \cite{zhou2021uhp} and
UHP-SOT++ \cite{zhou2022uhp}.  STRCF adds a temporal regularization term
to the objective function used for the regression of the feature map of
an object template in a DCF tracker. STRCF can effectively capture the
appearance change while being robust against abrupt errors. However, it
generates only rigid predictions and cannot recover from the tracking
loss. UHP-SOT enhances it with two modules: background motion modeling
and trajectory-based box prediction. The former models background
motion, conducts background motion compensation, and identifies the
salient motion of a moving object in a scene. It facilitates the
re-identification of the missing target after tracking loss. The latter
estimates the new location and shape of a tracked object based on its
past locations and shapes via linear prediction. The two modules can
collaborate together to estimate the box aspect ratio change to some
extent. UHP-SOT++ further improves the fusion strategy of different
modules and conducts more extensive experiments on the effectiveness of
each module on several tracking datasets. 

Although STRCF, UHP-SOT, and UHP-SOT++ boost the performance of classic
DCF trackers by a significant margin, their capability in flexible shape
estimation and object re-identification is still limited. This is
because they rely on the correlation between adjacent frames,
while an object template is vulnerable to shape deformation and
cumulative tracking errors in the long run. To improve the tracking
performance in long videos, GUSOT examines the shape estimation problem
and the object recovery problem furthermore. It exploits the spatial and
temporal correlation by considering foreground and background color
distributions. That is, colors in a search window are quantized into a
set of primary color keys. They are extracted across multiple frames
since they are robust against appearance change.  These salient color
keys can identify object/background locations with higher confidence.  A
low-cost graph-cut-based segmentation method can be used to provide the
object mask. GUSOT can accommodate flexible shape deformation to a
certain degree. 

All above-mentioned trackers model the object appearance from the global
view, i.e., using features of the whole object for the matching purpose.
They provide robust tracking results when the underlying object is
distinctive from background clutters without much deformation or
occlusion. For this reason, we adopt the global object-based correlator
in the first branch. The advanced version, GUSOT, is implemented in GOT. 

\subsection{Local Patch-based Correlator}\label{subsec:2nd}

The local patch-based correlator analyzes the temporal correlation
existing in parts of the tracked object. It is designed to handle object
deformations more effectively. It is formulated as a binary
classification problem. Given a local patch of size $8 \times 8$, the
binary classifier outputs its probability of being parts of the object
or the background. This is a novel contribution of this work. 

\subsubsection{Feature Extraction and Selection} 

\begin{figure}[htbp]
\centerline{\includegraphics[width=\linewidth]{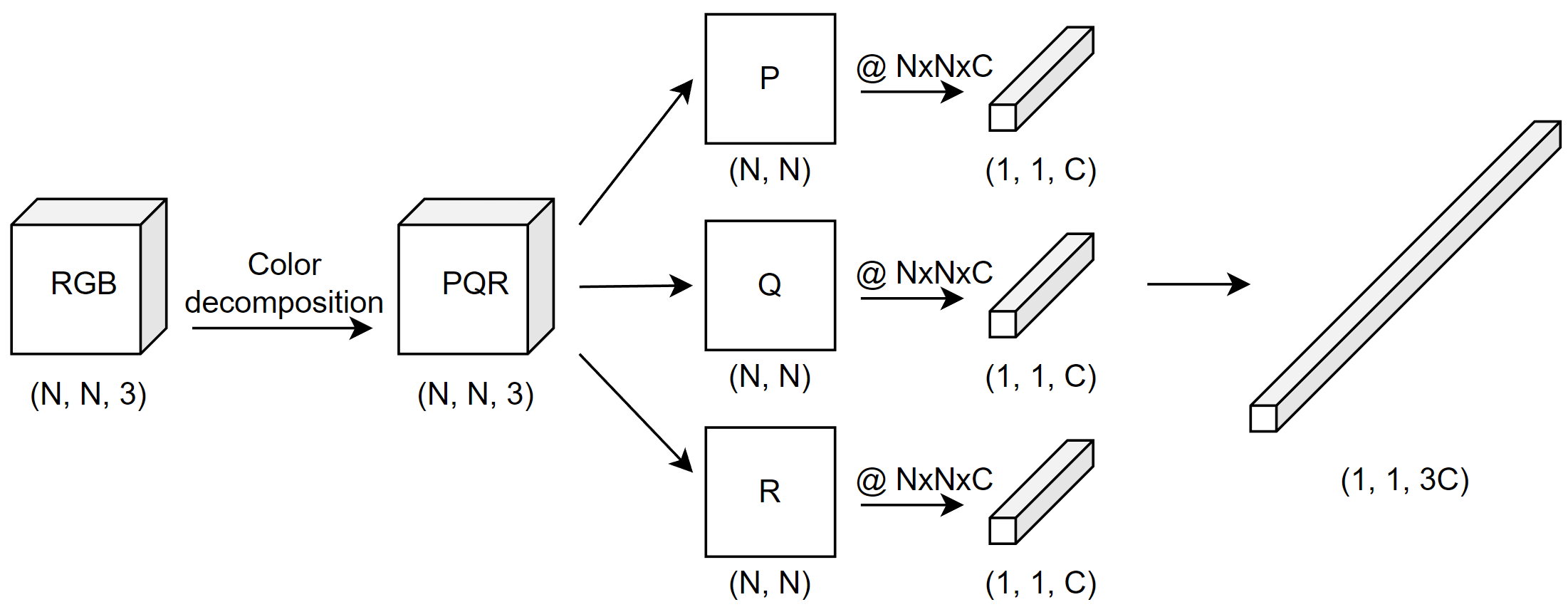}}
\caption{Channel-wise Saab transformation on color residuals of a patch of
size $N \times N$.} \label{fig:saab}
\end{figure}

The channel-wise Saab transform is an unsupervised representation
learning method proposed in \cite{chen2020pixelhop++}. It is slightly
modified and used to extract features of a patch here.  We decompose a
color input image into overlapping patches with a certain stride and
subtract the mean color of each patch to obtain its color residuals.
The mean color offers the average color of a patch.  The color
residuals are analyzed using the processing pipeline shown in
Fig.~\ref{fig:saab}, where the input consists of zero-mean RGB residual
channels.  We conduct the spectral principle component analysis (PCA) on
RGB residuals to get three decorrelated channels denoted by P, Q and R
channels.  For each of them, another spatial PCA is conducted to reduce
the feature dimension to $C$.  The final feature vector is formed by
concatenating of features of each color channel at each pixel. Note that
spectral and spatial PCA kernels are learned at the initial frame only
and shared among all remaining frames. Given the two PCA kernels, the
computation described above can be easily implemented by convolutional
layers of CNNs.  Besides the Saab features, handcrafted features such as
HOG and CN are also included for richer representation. Then, a feature
selection technique called discriminant feature test (DFT)
\cite{yang2022supervised} is adopted to select a subset of discriminant
features. The feature selection process is only conducted in the initial
frame. Once the features are selected, they are kept and shared among
later frames to reduce computational complexity.

\subsubsection{Patch Classification} 

If a patch is fully outside and inside the bounding box in the reference
frame, it is assigned ``0" and ``1", respectively. As for patches lying
on the box boundary, they are not used in training to avoid confusion.
Feature vectors and labels are used in an XGBoost classifier
\cite{chen2016xgboost}. In the experiment, we set the tree number and
the maximum tree depth to 40 and 4, respectively. These hyperparameters
are determined via offline cross validation on a set of training videos.
They can also be determined using samples in the initial frame.  The
predicted soft probability scores of patches in the search window of the
target frame form a heat map which is called the objectness score map.
Note that some patches inside the bounding box may belong to the
background rather than the object, leading to noisy labels. To alleviate
this problem, we adopt a two-stage training strategy.  The first-stage
classifier is trained using labels based on the patch location
inside/outside of the bounding box in the reference frame. It is applied
to patches in the target frame to produce soft probabilities.  Then, the
soft probabilities are binarized again to provide finetuned patch
labels. Due to the feature similarity between true background patches
and false foreground patches, their predicted soft labels should be
closer and, as a result, finetuned labels are more reliable than initial
labels. The second-stage classifier is trained using finetuned labels. 

\subsubsection{From Heat Map to Bounding Box} 

\begin{figure}[htbp]
\centerline{\includegraphics[width=\linewidth]{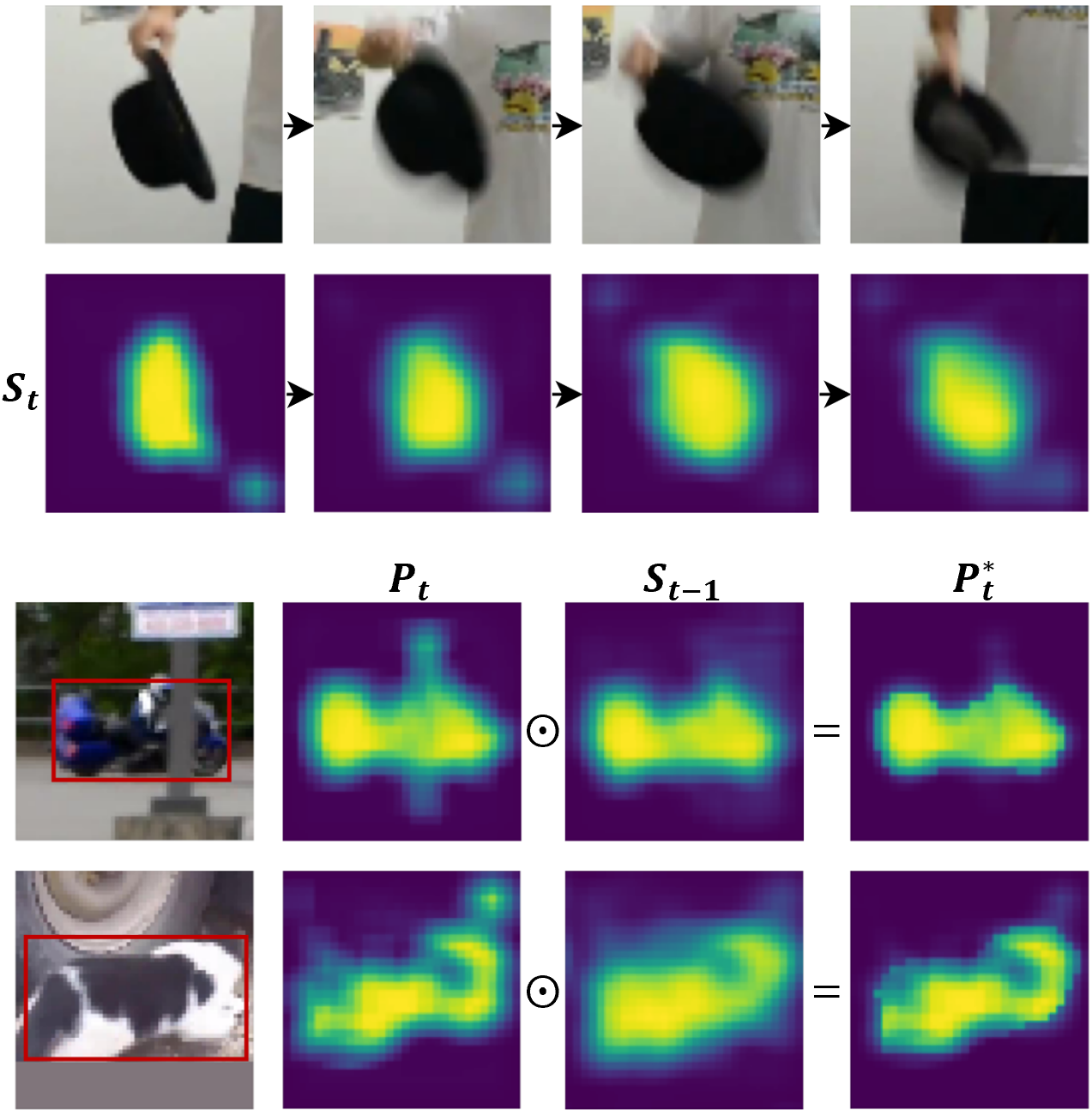}}
\caption{(Top) Visualization of the evolution of templates over time and
(bottom) visualization of the noise suppression effect in the raw
probability map based on Eq. (\ref{eq:shapetemp1}).}\label{fig:shapetemp}
\end{figure}

To obtain a rectangular bounding box, we binarize the heat map and draw
a tight enclosing box to obtain an objectness proposal. Due to noise
around the object boundary, direct usage of the heat map does not
yield stable box prediction. To overcome the problem, we smoothen the
heat map and use it to weigh the raw heat map for noise suppression. Let
$P_t \in \mathbf{R}^{H\times W}$, $S_{t-1}$, and $S_t$ denote the raw
probability map of frame $t$, the template of frame $t-1$ and the
updated template of frame $t$, respectively. Note that $S_{t-1}$ has
been registered to align with $P_t$ via circulant translation. The
processed heat map is expressed as
\begin{eqnarray}\label{eq:shapetemp1}
P_t^* = P_t \odot S_{t-1},
\end{eqnarray}
where $\odot$ is the element-wise multiplication for locations where
$S_{t-1}$ has the objectness score below 0.5. Then, $S_t$ is updated
by minimizing a cost function as follows:
\begin{eqnarray}\label{eq:shapetemp2}
S_t = \arg \min_{X}{\|X-P_t^*\|_{F}^2 + \mu \|X-S_{t-1}\|_{F}^2},
\end{eqnarray}
where parameter $\mu$ controls the tradeoff between the updating rate
and smoothness. Eq. (\ref{eq:shapetemp2}) is a regularized least-squares 
problem. It has the closed-form solution
\begin{equation}\label{eq:shapetemp3}
\begin{aligned}
S_t &= [P_t^*, \mu S_{t-1}] [I_H, \mu I_H]^\dag \\
& = [P_t^*, \mu S_{t-1}] ([1, \mu]\otimes I_H)^\dag  \\
& = [P_t^*, \mu S_{t-1}] ([1, \mu]^\dag \otimes I_H^\dag) \\
& = [P_t^*, \mu S_{t-1}] ([1, \mu]^\dag \otimes I_H) \\
& = [P_t^*, \mu S_{t-1}] ([\frac{1}{1+\mu^2}, \frac{\mu}{1+\mu^2}]^T \otimes I_H) \\
& = \frac{1}{1+\mu^2}P_t^* + \frac{\mu^2}{1+\mu^2}S_{t-1},
\end{aligned}
\end{equation}
where $\dag$, $\otimes$, and $I_H$ are the Moore–Penrose
pseudoinverse, the Kronecker product and the $H\times H$ identity
matrix, respectively.  We use several examples to visualize the
evolution of templates over time in Figure~\ref{fig:shapetemp}. 

\subsubsection{Classifier Update} 

\begin{figure}[htbp]
\centerline{\includegraphics[width=\linewidth]{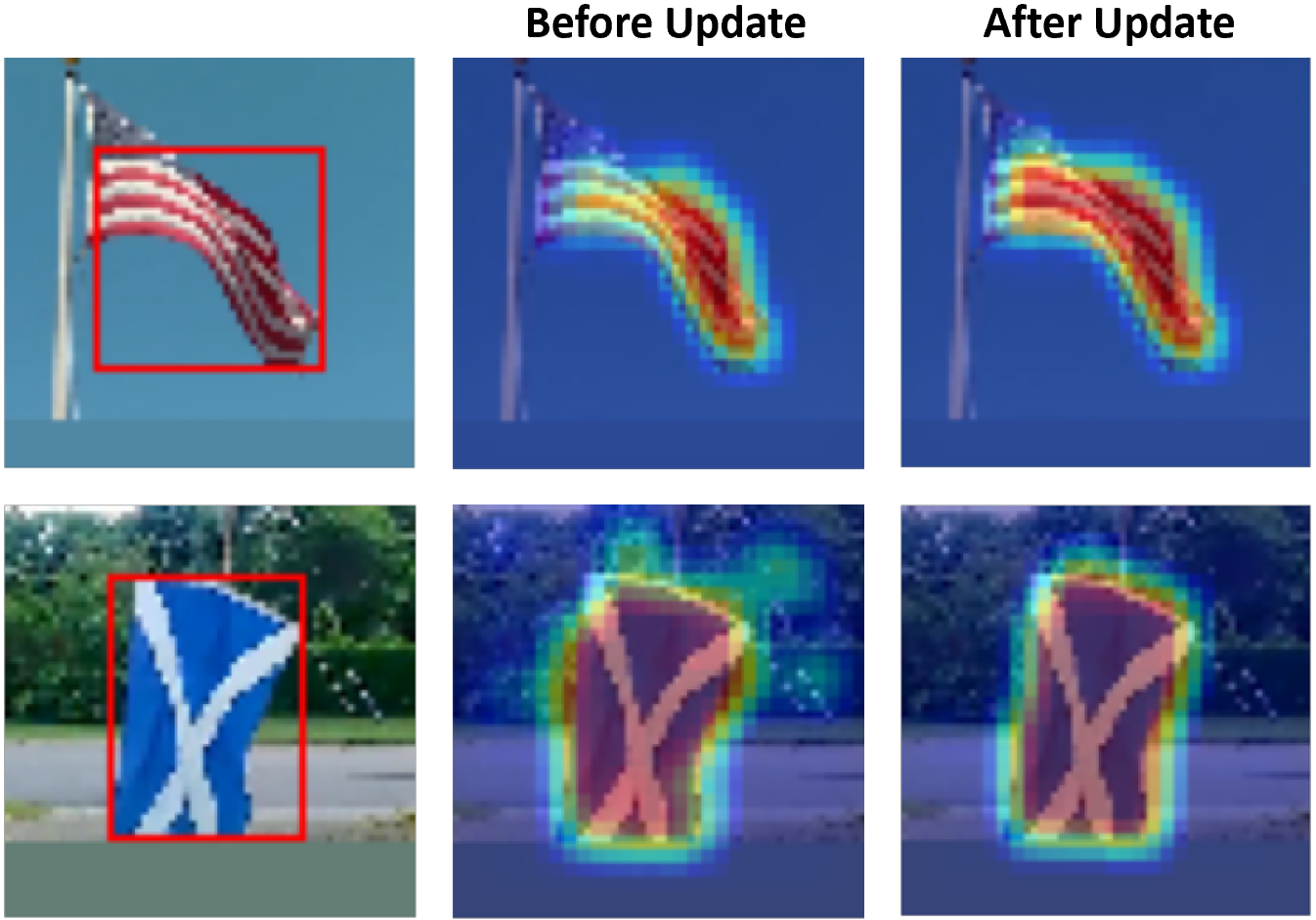}}
\caption{Proper updating helps maintain decent classification quality.}
\label{fig:clfupdate} \end{figure}

Since the object appearance may change over time, the classifier needs
to be updated to adapt to a new environment. The necessity of classifier
update can be observed based on the classification performance.  The
heat map is expected to span the object template reasonably well. If it
deviates too much from the object template, an update is needed.  As
shown in Fig.~\ref{fig:clfupdate}, regions of higher probability (marked
by warm colors) tend to shrink when there are new object appearances (in the
top example) or they may go out of the box when new background appears
(in the bottom example). Once one of such phenomena is observed, the
classifier should be retrained using samples from an earlier frame of
high confidence and those from the current frame. The retraining cost is
low because of the tiny size of the classifier. 

\subsection{Superpixel Segmentation} 

For the first and the second branches, we exploit temporal correlations
of the object and background across multiple frames.  In the third
branch, we consider spatial correlation in the target frame and perform
the unsupervised segmentation task.  Superpixel segmentation has been
widely studied for years. It offers a mature technique to generate a rough
segmentation mask. However, to group superpixels into a connected group,
an algorithm usually checks the appearance similarity and geometric
connectivity, which can be expensive. In our case, the heat map provides
a natural grouping guidance. When we overlay the heat map and superpixel
segments, each segment gets an averaged probability score. Then, we can
group segments by considering various probability thresholds and draw
multiple box proposals, as shown in Fig.~\ref{fig:system}.  They are
called superpixel proposals. 

\subsection{Fusion of Proposals}\label{subsec:integration}

There are three types of proposals in GOT: 1) the DCF proposal $x_{dcf}$ from the
global object-based correlator, 2) the objectness proposal $x_{obj}$ from
the local patch-based correlator, and 3) the superpixel proposals
$\chi_{spp}=\{x_{spp,i}|i=1,..,N\}$ from the superpixel segmentator. While
$\chi_{spp}$ contains multiple proposals by grouping different segments, the
most valuable one can be selected by evaluating the intersection-over-union (IoU)
as
\begin{equation}
x^*_{spp} = \arg \max_{x\in \chi_{spp}}{\text{IoU}(x,x_{dcf})+\lambda 
\text{IoU}(x,x_{obj})},
\end{equation}
where $\lambda$ is an adaptive weight calculated from $\text{IoU}
(x_{dcf},x_{obj})$. It lowers the contribution of poor heat maps.  In
the following, we first present two fusion strategies that combine
multiple proposals into one final prediction and then elaborate on
tracking quality control and object re-identification. 

\subsubsection{Two Fusion Strategies} 

According to the difficulty level of the tracking scenario, one final
box bounding is generated from these proposals with a simple or an
advanced fusion strategy. 

\textbf{Simple Fusion.} During an easy tracking period without obvious
challenges, multiple proposals tend to agree well with each other. Then,
we adopt a simple strategy based on IoU and probability values. 
The current tracking status is considered as easy if
\begin{eqnarray*}
\min_{x_i,x_j \in {x_{dcf},x_{obj},x_{spp}}} {\text{IoU}(x_i,x_j)} \geq \alpha,
\end{eqnarray*}
where $\alpha$ is the threshold to distinguish good and poor alignment.
Under this condition, we first fuse flexible proposals $x_{obj}$ and
$x_{spp}$ and then choose from the flexible proposal and the rigid
proposal $x_{dcf}$ via the following steps:
\begin{itemize}
\item Choose from $x_{obj}$ and $x_{spp}$ by finding $x_{df}^*= \arg
\max_{x}{\text{IoU}(x,x_{dcf})}$. 
\item Choose from $x_{df}^*$ and $x_{dcf}$. Stick to $x_{dcf}$ if it has
a larger averaged probability score inside the box and the size of 
$x_{def}^*$ changes too rapidly when compared with the previous prediction. 
\end{itemize}

\textbf{Advanced Fusion.} When multiple proposals differ a lot, it is
nontrivial to select the best one just using IoU or probability
distribution. Instead, we fuse the information from different
sources with the following optimization process. A rough foreground mask
$\mathbf{I^*}$ is derived by searching the optimal 0/1 label assignment
to pixels in the image. Let $x$ and $l_x$ denote the pixel location and
its label.  The mask, $\mathbf{I^*}$, can be estimated using the Markov Random
Field (MRF) optimization:
\begin{eqnarray}\label{eq:mrf}
\mathbf{I^*} = \arg \min_{\mathbf{I}}{\sum_{x}{\rho (x,l_x)} + 
\sum_{\{x,y\}\in \aleph}{w_{xy} \|l_x-l_y\|}},
\end{eqnarray}
where $\aleph$ is the four-connected neighborhood, the second term
assigns penalties to neighboring points that do not have the same labels, the
weight $w_{xy}$ is calculated from the color difference between $x$ and
$y$, and
\begin{eqnarray}\label{eq:mrfrou}
\rho (x,l_x) = -\log{p_{color}(l_x | x)} -\log{p_{obj}(l_x | x)}
\end{eqnarray}
treats the negative log-likelihood of $x$ being assigned to the
foreground color and that of $x$ being foreground in terms of
objectness. The former is modeled by the Gaussian mixtures while the
latter comes from the classification results.  The rough mask in Eq.
(\ref{eq:mrf}) takes color, objectness, and connectivity into account to
find the most likely label assignment.  While the solution could be
improved iteratively, we only run one iteration since the result is good
enough to serve as the rough mask.  Next, to fuse proposals, we select
the one that gets the highest IoU with the wrapping box of the rough
mask. If the advanced fusion fails, we go back to the simple fusion as
a backup. The overall fusion strategy that consists of both simple and
advanced fusion schemes is summarized in Algorithm \ref{algo:fusion}. 

\begin{algorithm}
\caption{Fusion of Multiple Proposals}\label{algo:fusion}
\begin{algorithmic} 
\renewcommand{\algorithmicrequire}{\textbf{Input:}}
\renewcommand{\algorithmicensure}{\textbf{Output:}}
\REQUIRE $x_{dcf}$, $x_{obj}$, $\chi_{spp}$, $\alpha$, $P_t^*$
\ENSURE final prediction $x_t$

\STATE $x_{spp} \leftarrow \arg \max_{x\in
\chi_{spp}}{\text{IoU}(x,x_{dcf})+\lambda \text{IoU}(x,x_{obj})}$ \STATE
$flag \leftarrow \{ \min_{x_i,x_j \in {x_{dcf},x_{obj},x_{spp}}}
{\text{IoU}(x_i,x_j)} \} \geq \alpha$ \IF {$flag$ is \FALSE} \STATE
generate box $x_{mrf}$ from MRF mask \IF {success} \RETURN $x_t
\leftarrow \arg \max_{x\in
x_{dcf},x_{obj},\chi_{spp}}{\text{IoU}(x,x_{mrf})}$ \ENDIF \ENDIF \STATE
$x_{df}^* \leftarrow \arg \max_{x\in
{x_{obj},x_{spp}}}{\text{IoU}(x,x_{dcf})}$ \STATE $S_{P_t^*}(x_{df}^*)
\leftarrow$ averaged probability inside $x_{df}^*$ \STATE
$S_{P_t^*}(x_{dcf}) \leftarrow$ averaged probability inside $x_{dcf}$
\IF {$x_{df}^*$ is stable \OR $S_{P_t^*}(x_{df}^*) >
S_{P_t^*}(x_{dcf})$} \RETURN $x_t \leftarrow x_{df}^*$ \ENDIF \RETURN
$x_t \leftarrow x_{dcf}$

\end{algorithmic}
\end{algorithm}

\subsubsection{Tracking Quality Control} 

\begin{figure}[htbp]
\centerline{\includegraphics[width=\linewidth]{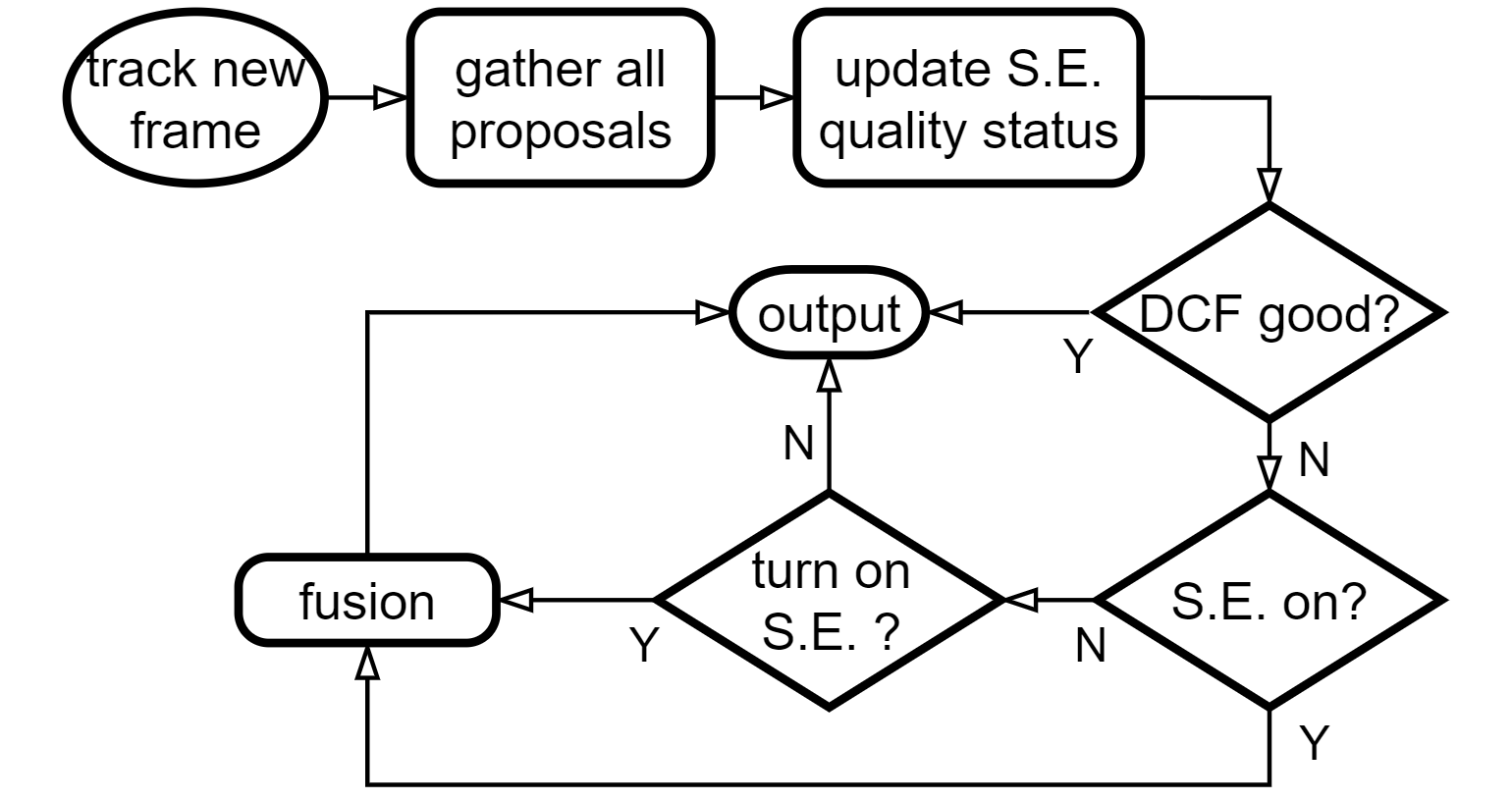}}
\caption{Management of different tools for tracking, where S.E. denotes the shape
estimation function provided by the second branch.} \label{fig:flowchart_fusion}
\end{figure}

For rigid objects with rigid motion, the global object-based correlator
can provide a fairly good prediction. The template matching similarity
score of DCF is usually high. The local patch-based correlator usually
helps in the face of challenges such as background clutters and occlusions.
However, due to the complicated nature of local patch classification,
its proposal may be noisy.  Detection and removal of noisy proposals
in the second branch are important to good tracking performance in
general. To solve this issue, we monitor the quality of the heat map and
may discard noisy proposals until classification gets stable again.  The
flowchart of tracking quality control is depicted in
Fig.~\ref{fig:flowchart_fusion}. After the heat map is obtained, we
check whether the high probability region is too small or too large and
whether it contains several unconnected blobs. All of them indicate that
the local patch-based correlator is not stable. Thus, its proposal is
discarded. Once the problem is resolved, the heat map becomes stable,
and the objectness proposal shall have small variations in height and
width. Then, we can turn on the shape estimation functionality (i.e.,
the local patch-based correlator in the second branch) and conduct
the fusion of all proposals. 

\begin{figure*}[htbp]
\centerline{\includegraphics[width=\textwidth]{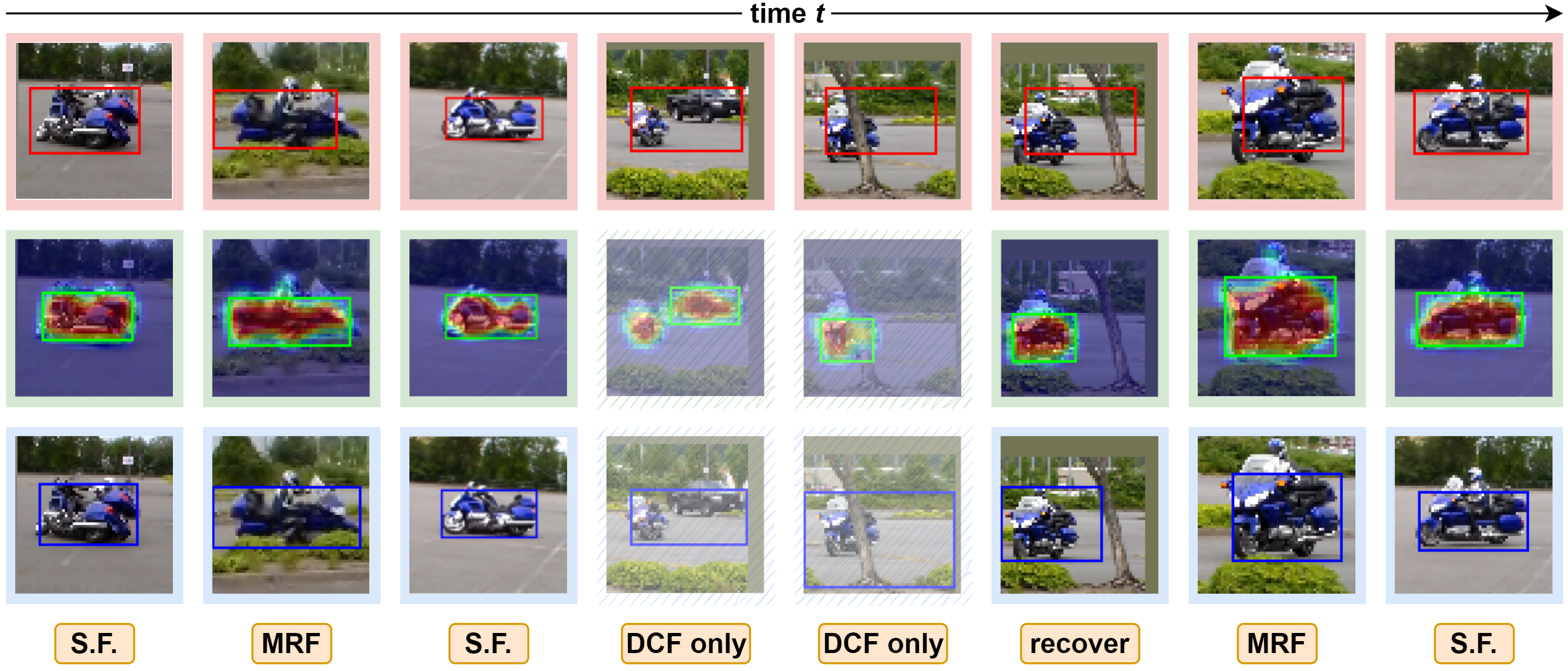}}
\caption{The fusion strategy (given in the fourth row, where S.F. stands
for simple fusion) changes with tracking dynamics over time. The DCF
proposal, the objectness proposal, and the superpixel proposal are given
in the first, second, and third rows, respectively.  The sequence is
\textit{motorcycle-9} from LaSOT and the object is the motorbike.}
\label{fig:dynamic}
\end{figure*}

An exemplary sequence is illustrated in Fig.~\ref{fig:dynamic}. In the
beginning, the tracking process is smooth and the simple fusion is
sufficient.  Then, when some challenges appear and multiple proposed
boxes do not align, we turn to the advanced MRF fusion. When there is
severe background clutter or occlusion that confuses the classifier in
the second branch, the DCF proposal is adopted directly until the
turbulence goes away. Then, the process repeats until the end of the
video. 

\subsubsection{Object Re-identification} 

Besides shape estimation, the objectness proposal can be used for object
re-identification after tracking loss. Given the current DCF proposal
and the motion proposal that covers the most motion flow and possibly
contains the lost foreground object, GUSOT selects one of the two via
trajectory stability and color/template similarity. However, GUSOT is
not general enough to cover all cases. The objectness score provides an
extra view of the appearance similarity with timely updated object
information, and it helps recover the object quickly. 

Given a candidate box proposal $x$ with center $x_{ct}$, its averaged
objectness score $S_{obj}(x)$ inside the box, the feature representation
$f(x)$ of the region, and the DCF template $f_{t-1}$, the scoring
function for this candidate can be calculated as
\begin{equation}
S(x) = \beta_1 \langle f(x),f_{t-1} \rangle + \beta_2 S_{obj}(x) 
- \beta_3 \|x_{ct}-\hat{x}_{t,ct}\|^2 , 
\end{equation}
where $\hat{x}_t$ is the linear prediction of the box center based on
past predictions, $\beta_1$ and $\beta_3$ are positive constants to
adjust the magnitudes of all terms to the same level, $\beta_2$ is 
a positive adaptive weight 
that assigns lower contributions for poorer probability maps. An ideal
candidate box is expected to have high template similarity, high
objectness score, and with a small translation from the last prediction.
Then, we can choose from the DCF proposal and the motion proposal by
selecting the one with a higher evaluation score to re-identify the lost
object.

\section{Experiments}\label{sec:experiments}

\subsection{Experimental Setup}

\subsubsection{Performance Metrics}

The one-pass-evaluation (OPE) protocol is adopted for all trackers in
performance benchmarking unless specified otherwise.  The metrics for
tracking accuracy include: the distance precision (DP) and the
area-under-curve (AUC).  DP measures the center precision at the
20-pixel threshold to rank different trackers, and AUC is calculated
using the overlap precision curve. For model complexity, we consider two
metrics: the model size and the computational complexity required to
predict the target object box from the reference one on average. The
latter is also called the inference complexity per frame.  The model
size is the number of model parameters.  The inference complexity per
frame is measured by the number of (multiplication or add) floating
point operations (flops). 

\subsubsection{Benchmarking Object Trackers}

We compare GOT with the following four categories of trackers. 
\begin{itemize}
\item supervised lightweight DL trackers: LightTrack \cite{yan2021lighttrack},
DSTfc \cite{shen2021distilled}, and FEAR-XS \cite{borsuk2022fear}.
\item supervised DL trackers: SiamFC \cite{bertinetto2016fully},
ECO \cite{danelljan2017eco}, and SiamRPN \cite{li2018high}.
\item unsupervised DL trackers: LUDT \cite{wang2021unsupervised},
ResPUL \cite{wu2021progressive}, USOT \cite{zheng2021learning},
and ULAST \cite{shen2022unsupervised}.
\item unsupervised DCF trackers: KCF \cite{henriques2014high},
SRDCF \cite{danelljan2015learning}, and STRCF \cite{li2018learning}.
\end{itemize}
For unsupervised DL trackers, we use the models trained from scratch for
comparison if they are available. 

\subsubsection{Tracking Datasets}

We conduct performance evaluations of various trackers on four datasets.
\begin{itemize}
\item OTB2015 \cite{7001050}. It contains 100 videos with an average
length of 598 frames. The dataset was released in 2015. Many video
sequences are of lower resolution. 
\item VOT2016 \cite{hadfield2016visual}. It contains 60 video sequences
with an average length of 358 frames. It has a significant overlap with
the OTB2015 dataset. One of its purposes is to detect the frequency of
tracking failures. Different from the OPE protocol, once a failure is detected, the baseline experiment
re-initializes the tracker.
\item TrackingNet \cite{muller2018trackingnet}. It is a large-scale
dataset for object tracking in the wild. Its testing set consists of 511
videos with an average length of 442 frames. 
\item LaSOT \cite{fan2019lasot}. It is the largest single object
tracking dataset by far. It has 280 long testing videos with 685K frames
in total. The average video length is 2000+ frames. Thus, it serves as
an important benchmark for measuring long-term tracking performance. 
\end{itemize}

\begin{table*}[htbp]
\caption{Comparison of tracking accuracy and model complexity of
representative trackers of four categories on four tracking datasets.
Some numbers for model complexity are rough estimations due to the lack of detailed
description of network structures and/or public domain codes.
Furthermore, the complexity of some algorithms is related to built-in
implementation and hardware.  \textbf{OPT} and \textbf{UT} are
abbreviations for offline pre-training and unsupervised trackers,
respectively. The top 3 runners among all unsupervised trackers (i.e.,
those in the last two categories) are highlighted in red, green, and
blue, respectively.}\label{tab:comp}
\begin{center}
\renewcommand*{\arraystretch}{1.3}
\begin{tabular}{ccccccccccc} \hline
& & & & \multicolumn{2}{c}{OTB2015}& VOT16 & \multicolumn{2}{c}{TrackingNet} & \multicolumn{2}{c}{LaSOT} \\
Tracker & OPT/UT & Flops ($\downarrow$) &  No. of Params.($\downarrow$)& DP ($\uparrow$) & AUC ($\uparrow$)&EAO($\uparrow$) & DP($\uparrow$) & AUC($\uparrow$) & DP ($\uparrow$) & AUC ($\uparrow$) \\ \hline
LightTrack\cite{yan2021lighttrack} & \checkmark/$\times$ & 530 M (9X) & 1.97 M (896X) & - &66.2&-&69.5&72.5& 53.7 & 53.8 \\
DSTfc\cite{shen2021distilled} & \checkmark/$\times$ & 1.23 G (21X) & 0.54 M (246X) & 76.1 &57.3&-&51.2&56.2& - & 34.0 \\
FEAR-XS\cite{borsuk2022fear} & \checkmark/$\times$ & 478 M (8X) & 1.37 M (623X) & - &-&-&-&-& 54.5 & 53.5 \\ \hline
SiamFC\cite{bertinetto2016fully} & \checkmark/$\times$ & 2.7 G (47X) & 2.3 M (1046X) & 77.1 &58.2&23.5&53.3&57.1& 33.9 & 33.6 \\
ECO\cite{danelljan2017eco} & \checkmark/$\times$ & 1.82 G (31X) & 95 M (43201X) &90.0&68.6&37.5&48.9&56.1& 30.1 & 32.4 \\
SiamRPN\cite{li2018high} & \checkmark/$\times$ & 9.23 G (159X) & 22.63 M (10291X) &85.1&63.7&34.4&-&-& 38.0 & 41.1\\ \hline
LUDT\cite{wang2021unsupervised} & \checkmark/\checkmark & -  & -  & 76.9 & 60.2 & 23.2 & 46.9 & 54.3 & - & 26.2\\
ResPUL\cite{wu2021progressive} & \checkmark/\checkmark & 2.65 G (46X) & 1.445 M (657X) & - & 58.4 & 26.3 & \textcolor{blue}{48.5} & \textcolor{blue}{54.6} & - & -\\
USOT\cite{zheng2021learning} & \checkmark/\checkmark & $>$14 G (241X) & 29.4 M (13370X) & 79.8 & 58.5 & \textcolor{red}{35.1} & \textcolor{red}{55.1} & \textcolor{red}{59.9} & \textcolor{blue}{32.3} & \textcolor{blue}{33.7}\\
ULAST\cite{shen2022unsupervised} & \checkmark/\checkmark & $\gtrapprox$50 G (862X) & $\gtrapprox$50 M (22738X) & \textcolor{blue}{81.1}& \textcolor{blue}{61.0} &-&-&-& \textcolor{red}{40.7} & \textcolor{red}{43.3}\\ \hline
KCF\cite{henriques2014high} & $\times$/\checkmark & - & - & 69.6 & 48.5 & 19.2 & 41.9 & 44.7 & 16.6 & 17.8\\
SRDCF\cite{danelljan2015learning} & $\times$/\checkmark & -  & -  & 78.1 & 59.3 & 24.7 & 45.5 & 52.1 & 21.9 & 24.5\\
STRCF\cite{li2018learning} & $\times$/\checkmark & -  & -  &\textcolor{ForestGreen}{86.6}&\textcolor{red}{65.8}&\textcolor{ForestGreen}{27.9}&-&-& 29.8 & 30.8\\
GOT (Ours) & $\times$/\checkmark & $\approx$\textbf{58 M} (1X) & \textbf{2199} (1X) & \textcolor{red}{87.6} & \textcolor{ForestGreen}{65.4} & \textcolor{blue}{26.8} & \textcolor{ForestGreen}{52.6} & \textcolor{ForestGreen}{56.3} & \textcolor{ForestGreen}{38.8} & \textcolor{ForestGreen}{38.5}\\ \hline
\end{tabular}
\end{center}
\end{table*}


\subsubsection{Implementation Details}

In the implementation, each region of interest is warped into a
$60\times60$ patch with an object that takes around $32\times32$ pixels.
The XGBoost classifier in the local correlator has 40 trees with the
maximum depth set to 4.  Parameters $\alpha=0.7$ and $\mu=5$ are used in
the fuser.  The first branch (i.e., the global correlator), the
combined second and third branches (i.e., the local correlator and the
superpixel segmentator), and the fuser runs at 15 FPS, 5 FPS, and
15 FPS on one Intel(R) Core(TM) i5-9400F CPU, respectively. The speed
can be further improved via code optimization and parallel programming. 

\subsection{Performance Evaluation}

We compare the performance of GOT with four categories of trackers on
four datasets in Table \ref{tab:comp}. Trackers are grouped based on
their categories. From top to down, they are supervised lightweight DL
trackers, supervised DL trackers, unsupervised DL trackers, and
unsupervised DCF trackers, respectively. Our proposed GOT belongs to the
last category. We have the following observations.

\textbf{OTB2015}. GOT has the best performance in DP and the second best
performance in AUC among unsupervised trackers on this dataset. One explanation is that DL trackers
are trained on high-resolution videos and they do not generalize well to
low-resolution videos. GOT is robust against different resolutions since
its HoG and CN features are stable to ensure a higher successful rate of
the template matching idea. 

\textbf{VOT2016}. It adopts the expected average overlap (EAO) metric to
evaluate the overall performance of a tracker. EAO considers both
accuracy and robustness. The observation on this dataset reveals that the advantages of the local correlator
branch are not obvious on very short videos since the tracker gets
corrected automatically if its IoU is lower than a threshold. Yet, GOT
still ranks third among unsupervised trackers (i.e., the last two
categories) with a tiny model (of 2.2K parameters) and much lower
inference complexity by 3 to 5 orders of magnitude. 

\textbf{TrackingNet}. The ground-truth box is provided for the first
frame only. The performance of GOT is evaluated by an online server. GOT
ranks second among unsupervised trackers. Its performance is also
comparable with almost all supervised DL trackers (except LightTrack).

\textbf{LaSOT}. GOT ranks second among unsupervised trackers. It
even has better performance than some supervised trackers such as DSTfc
while maintaining a much smaller model size and lower computational
complexity. 

\subsection{Comparison Among Lightweight Trackers}

We compare the design methods and training costs of GOT and three
lightweight DL trackers in Table \ref{tab:lightweight}.  The lightweight
DL trackers conduct the neural architecture search (NAS) or model
distillation/optimization to reduce the model size and inference
complexity.  As shown in Table \ref{tab:comp}, LightTrack achieves even
higher tracking accuracy than large models. Besides NAS, FEAR-XS
\cite{borsuk2022fear} adopts several special tools such as depth-wise
separable convolutions and increases the number of object templates to
lower complexity while maintaining high accuracy. Although DSTfc has the
smallest model size among the three, its model size is still larger than
that of GOT by two orders of magnitude. Furthermore, the tracking
performance of GOT is better than that of DSTfc in three datasets.  As
to the training cost, all three lightweight DL trackers need
pre-training on millions of labeled frames while GOT does not require
any as shown in the last column of Table \ref{tab:comp}.  The
superiority of LightTrack in accuracy does have a cost, including long
pre-training, architecture search and fine-tuning. Finally, it is worth
mentioning that GOT is more transparent in its design. Thus, its source
of tracking errors can be explained and addressed. In contrast, the
failure of lightweight DL trackers is difficult to analyze. It could be
overcome by adding more pre-training samples and repeating the whole
design process one more time. 

\begin{table}[htbp]
\caption{Comparison of design methods and pre-training costs of 
GOT and three lightweight DL trackers.}\label{tab:lightweight}
\begin{center}
\resizebox*{\linewidth}{!}{
\begin{tabular}{ccc}\hline
Trackers & Design Methods & \# of Pre-Training Boxes\\ \hline
LightTrack\cite{yan2021lighttrack} & NAS & $\approx10$M \\
DSTfc\cite{shen2021distilled} & NAS, model distillation &  $\approx2$M\\ 
FEAR-XS \cite{borsuk2022fear}& NAS, network optimization &  $\approx13$M\\
GOT (Ours) & fusion of 3 branches & 0 \\ \hline
\end{tabular}
}
\end{center}
\end{table}


\subsection{Long-term Tracking Capability}

To examine GOT's capability in long-term tracking, we test it on the
test set of the OxUva dataset \cite{valmadre2018long}. OxUva contains
166 long test videos under the tracking-in-the-wild setting.  The object
to be tracked disappears from the field of view in around one half of
video frames. Trackers need to report whether the object is present or
absent and give the object box when it is present. The ground-truth
labels are hidden, and the tracking results are evaluated on a
competition server. Since the competition is not maintained any longer,
we cannot submit our results for official evaluation. For this reason,
we use predictions from the leading tracker LTMU \cite{dai2020high} as
the pseudo labels for performance evaluation below. 
There are three major evaluation metrics: the true positive rate (TPR), 
the true negative rate (TNR) and the maximum geometric mean (MaxGM), 
which is calculated as
\begin{eqnarray} 
\text{MaxGM} = \max_{0\leq p\leq 1}{\sqrt{(1-p)\cdot\text{TPR} 
\cdot((1-p) \cdot\text{TNR}+p)}},
\end{eqnarray} 
where TPR stands for the fraction of presented objects that are
predicted as present and located with a tight bounding box, and TNR
represents the fraction of absent objects that are correctly reported as
absent. 

\begin{figure}[htbp]
\centerline{\includegraphics[width=0.7\linewidth]{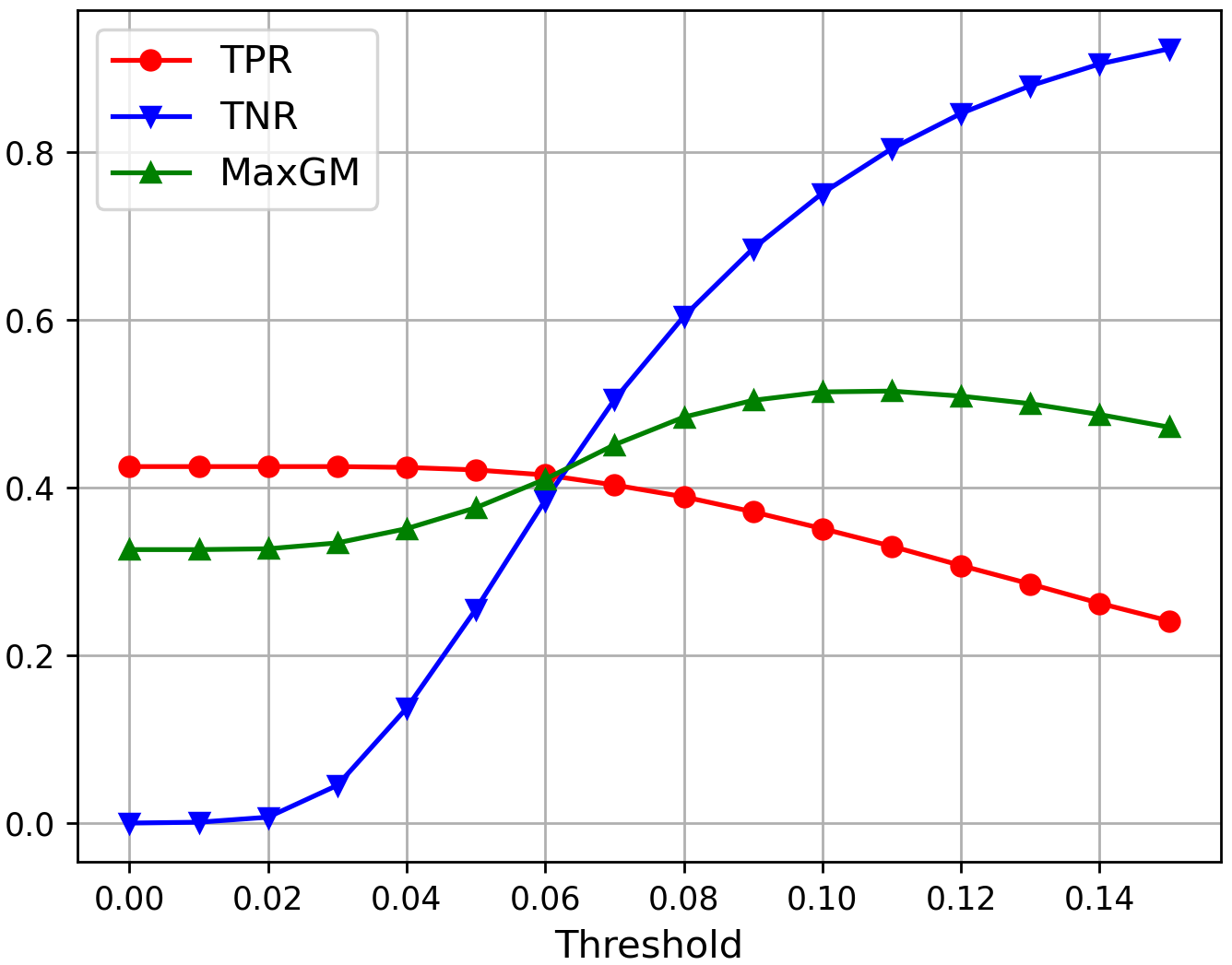}}
\caption{The TPR, TNP and MaxGM values of GOT$^*$ at different present/absent
threshold values against the OxUva dataset.}\label{fig:ox_ours_th}
\end{figure}

We compare the above performance metrics of GOT against three lightweight long-term
trackers,
KCF \cite{henriques2014high} (the long-term version), TLD \cite{kalal2011tracking} and FuCoLoT
\cite{lukevzivc2019fucolot}, in Table \ref{tab:ox}. TLD and FuCoLoT are
equipped with a re-detection mechanism to find the object after loss.
We see that GOT achieves the highest TPR because of its accurate box
predictions.  FuCoLoT and GOT do not provide present/absent predictions
so their TNR values are zero. In GOT$^*$, the object is claimed to be
absent if the similarity score of template matching is lower than a
threshold, which is set to $0.1$ in the experiment.  Then, KCF and
GOT$^*$ have comparable performance in TNR.  Finally, GOT$^*$ has the
best performance in MaxGM.  The above design indicates that the
similarity score in GOT is a simple yet effective indicator of the
object status. The effect of various threshold values on GOT$^*$ is
illustrated in Fig.~\ref{fig:ox_ours_th}. As the threshold grows from
$0$ to $0.15$, its TPR decreases slowly while its TNR increases quickly.
The optimal threshold for the MaxGM metric is around $0.1$. 

\begin{table}[htbp]
\caption{Performance comparison of GOT and GOT$^*$ against three lightweight long-term
trackers, KCF (the long-term version), TLD and
FuCoLoT, on the OxUvA dataset, where the best performance is shown in
red. KCF and TLD are implemented in OpenCV.}\label{tab:ox}
\begin{center}
\resizebox*{\linewidth}{!}{
\begin{tabular}{cccccc}\hline
     & KCF & TLD & FuCoLoT & GOT & GOT$^*$\\ \hline
TPR ($\uparrow$) & 0.165 & 0.142 & 0.353 & \textcolor{red}{0.425} & 0.351 \\
TNR ($\uparrow$) & \textcolor{red}{0.872} & 0.095 & 0 & 0 & 0.751 \\ 
MaxGM ($\uparrow$) & 0.380 & 0.198 & 0.297 & 0.326 & \textcolor{red}{0.514}\\ \hline
\end{tabular}
}
\end{center}
\end{table}

\subsection{Attribute-based Performance Evaluation} 

\begin{figure}[htbp]
\centerline{\includegraphics[width=\linewidth]{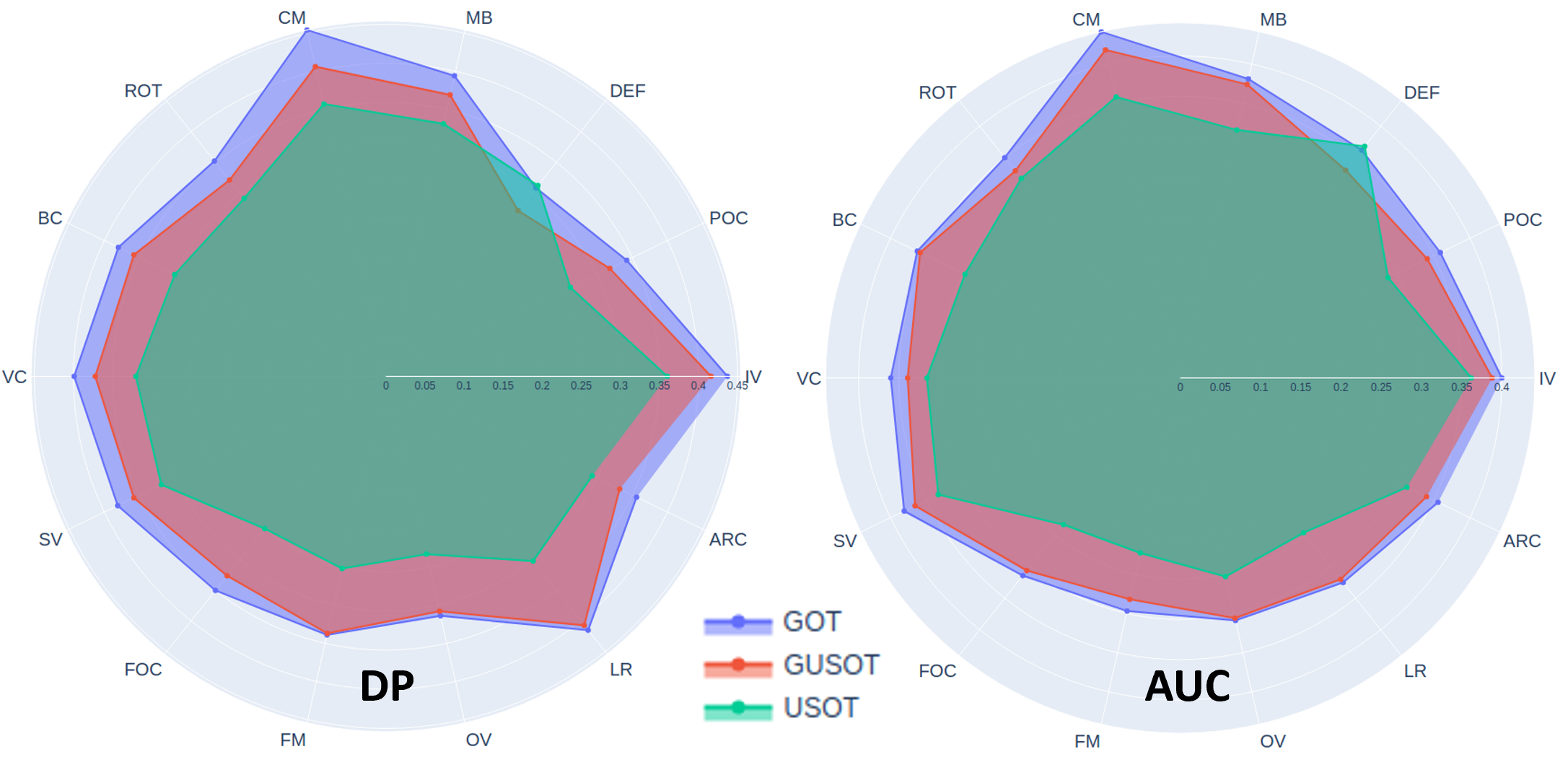}}
\caption{Attribute-based evaluation of GOT, GUSOT and USOT on
LaSOT in terms of DP and AUC, where attributes of interest include the
aspect ratio change (ARC), background clutter (BC), camera motion (CM),
deformation (DEF), fast motion (FM), full occlusion (FOC), illumination
variation (IV), low resolution (LR), motion blur (MB), occlusion (OCC),
out-of-view (OV), partial occlusion (POC), rotation (ROT), scale
variation (SV) and viewpoint change (VC).} \label{fig:ablation1}
\end{figure}

\begin{figure*}[htbp]
\centerline{\includegraphics[width=\textwidth]{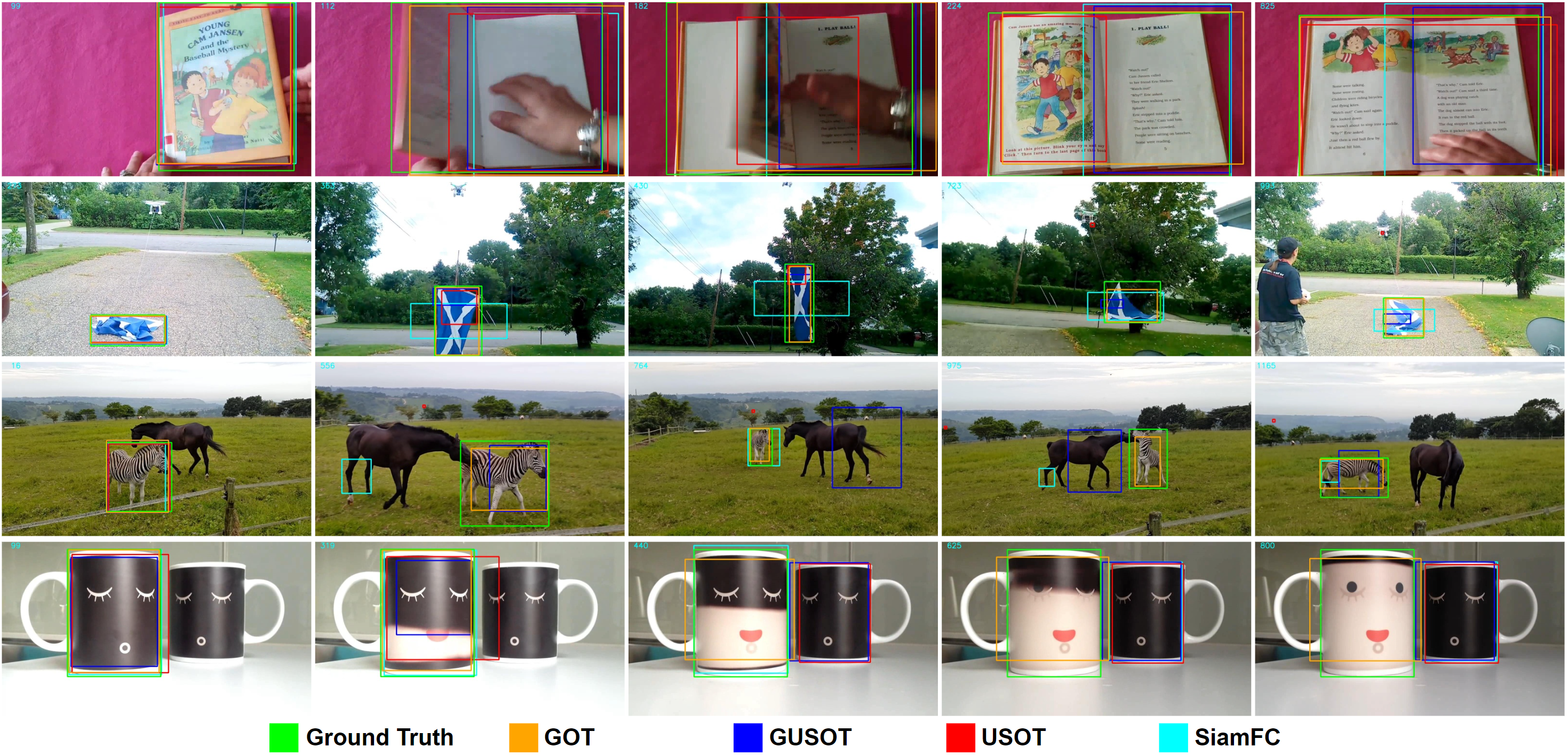}}
\caption{Comparison of the tracked object boxes of GOT, GUSOT, USOT, and
SiamFC for four video sequences from LaSOT (from top to bottom: book,
flag, zebra, and cups).  The initial appearances
are given in the first (i.e., leftmost) column. The tracking results for
four representative frames are illustrated.} \label{fig:samples}
\end{figure*}

To shed light on the strengths and weaknesses of GOT, we conduct the
attribute-based study among GOT, the GUSOT baseline, and USOT, which is
an unsupervised DL tracker, on the LaSOT dataset.  DPs and AUCs with
respect to different challenging attributes are presented in
Fig.~\ref{fig:ablation1}.  GUSOT outperforms USOT in all aspects
except deformation (DEF) due to its limited segmentation-based shape
adaptation capability. With the help of the local correlator branch and
the powerful fuser, GOT performs better than GUSOT and has
comparable performance with USOT, which is equipped with a box
regression network, in DEF.  For the same reason, GOT outperforms GUSOT
in other DEF-related attributes such as viewpoint change (VC), rotation
(ROT), and aspect ratio change (ARC).  Another improvement lies is
camera motion (CM), where the local correlator contributes to better
object re-identification.  GOT has the least improvement in low
resolution (LR). It appears that both the local correlator and the
superpixel segmentation cannot help the GUSOT baseline much in this
case. Lower video resolutions make the local features (say, around the
boundaries) less distinguishable from each other. 

As discussed above, GOT adapts to the new appearance and shape well
against the DEF challenge.  Representative frames of four video
sequences from LaSOT are illustrated in Fig.~\ref{fig:samples} as
supporting evidence. All tracked objects have severe deformations.  For
the first sequence of turning book pages, GOT covers the whole book
correctly.  For the second sequence of a flag, GOT can track the flag
accurately.  For the third sequence of a zebra, object re-identification
helps GOT relocate the prediction to the correct place once the object
is free from occlusion.  For the fourth sequence of two cups, GOT is
robust against background clutters.  In contrast, other trackers fail to
catch the new appearance or completely lose the object. 

\subsection{Insights into GOT's New Ingredients}

GOT has two new ingredients: 1) the local correlator in the 2nd branch
and 2) the fuser to combine the outputs from all three branches. We
provide further insights into them below. 

\subsubsection{Local Correlator} 

We compare the performance of GOT under different settings on the
LaSOT dataset in Table~\ref{tab:ablation_clf}. The settings include:
\begin{itemize}
\item With or without the local correlator branch;
\item With or without classifier update;
\item With or without object re-identification.
\end{itemize}
We see from the table that a ``plain" local correlator already achieves
a substantial improvement in DP. Classifier update and/or object
re-identification improve more in AUC but less in DP. This is because
deformation tends to happen around object boundaries and the change of
the object center is relatively slow.  On the other hand, the addition
of classifier update and object re-identification helps improve the
quality of the objectness map for better shape estimation. In addition,
the improvement from object re-identification indicates the frequent
object loss in long videos and the effective contribution of the objectness
score.  Both classifier update and object re-identification are needed
to achieve the best performance. 

\begin{table}[htbp]
\caption{Performance comparison of GOT under different settings on the
LaSOT dataset, where the best performance is shown in red. The ablation
study includes: 1) with or without the local correlator branch; 2) with
or without classifier update; 3) with or without object
re-identification.}\label{tab:ablation_clf}
\begin{center}
\resizebox*{\linewidth}{!}{
\begin{tabular}{ccccc}\hline
L.C.B. & Clf. update  & Re-idf. & DP ($\uparrow$) & AUC ($\uparrow$) \\ \hline
 &  &  & 36.1 & 36.8  \\
\checkmark &  &  & 38.0 & 37.5 \\ 
\checkmark & \checkmark &  & 38.2 & 38.0 \\ 
\checkmark &  & \checkmark & 38.2 & 37.9 \\ 
\checkmark & \checkmark & \checkmark & \textcolor{red}{38.8} & \textcolor{red}{38.5} \\ \hline
\end{tabular}
}
\end{center}
\end{table}

To study the necessity of quality checking and maintenance in the
classification system of the local correlator branch, we compare the
tracking accuracy of three settings in Table \ref{tab:ablation_clf_},
where object re-identification is turned off in all settings.  Without
the shape estimation on-off scheme, the tracker simply stops the shape
estimation function after failures. The performance drops, which reveals
the frequent occurrence of challenges even in the early/middle stage of
videos and the importance of quality checking. Noise suppression helps
boost the tracking accuracy furthermore since it alleviates abrupt box
changes due to noise around the object border. 

\begin{table}[htbp]
\caption{Ablation study of the classification system in the local
correlator branch in GOT on LaSOT under three settings, where the best
performance is shown in red.} \label{tab:ablation_clf_}
\begin{center}
\resizebox*{\linewidth}{!}{
\begin{tabular}{cccc}\hline
 Shape Estimation On-Off & Noise Suppression & DP ($\uparrow$) & AUC ($\uparrow$) \\ \hline
\checkmark &  & 37.6 & 37.8  \\
 & \checkmark & 36.5 & 36.5 \\ 
\checkmark & \checkmark & \textcolor{red}{38.2} & \textcolor{red}{38.0} \\ \hline
\end{tabular}
}
\end{center}
\end{table}

\subsubsection{Fuser} 

\begin{figure}[htbp]
\centerline{\includegraphics[width=\linewidth]{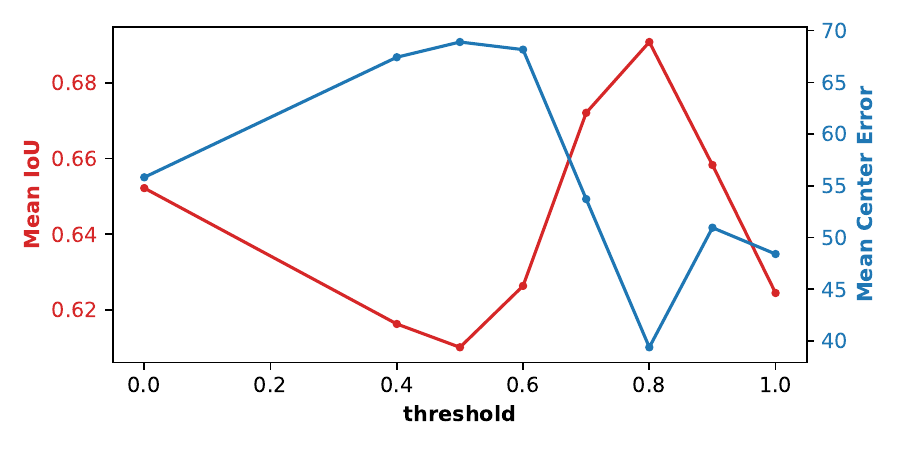}}
\caption{Mean IoU (the higher the better) and center error (the lower
the better) on the selected subset with different fusion thresholds. The
subset from LaSOT includes \textit{book-10}, \textit{bus-2},
\textit{cat-1}, \textit{crocodile-14}, \textit{flag-5}, \textit{flag-9},
\textit{gorilla-6}, \textit{person-1}, \textit{squirrel-19},
\textit{mouse-17}.} \label{fig:ab_th_lasot}
\end{figure}

The threshold parameter, $\alpha$, in Algorithm \ref{algo:fusion} is
used to choose between the simple fusion or the MRF fusion. To study its
sensitivity, we select a subset of 10 sequences from LaSOT and turn on
shape estimation in most frames. The mean IoU and the center error
between the ground truth and the prediction change are plotted as
functions of the threshold value in Fig.~\ref{fig:ab_th_lasot}.  The
optimal threshold range is between 0.7 and 0.9 since it has lower center
errors and higher IoUs. Choosing a lower threshold means that we conduct
the simple fusion.  This is consistent with the proposed fusion
strategy. That is, the simple fusion should be only used when different
proposals are close to each other. Pure simple and MRF fusion strategies
have their own weaknesses such as the limited selection ability in the
simple fusion and errors around boundaries in the MRF fusion. Proper
collaboration between them can boost the performance. 

\section{Discussion on GOT's Limitations}

\begin{figure}[htbp]
\centerline{\includegraphics[width=\linewidth]{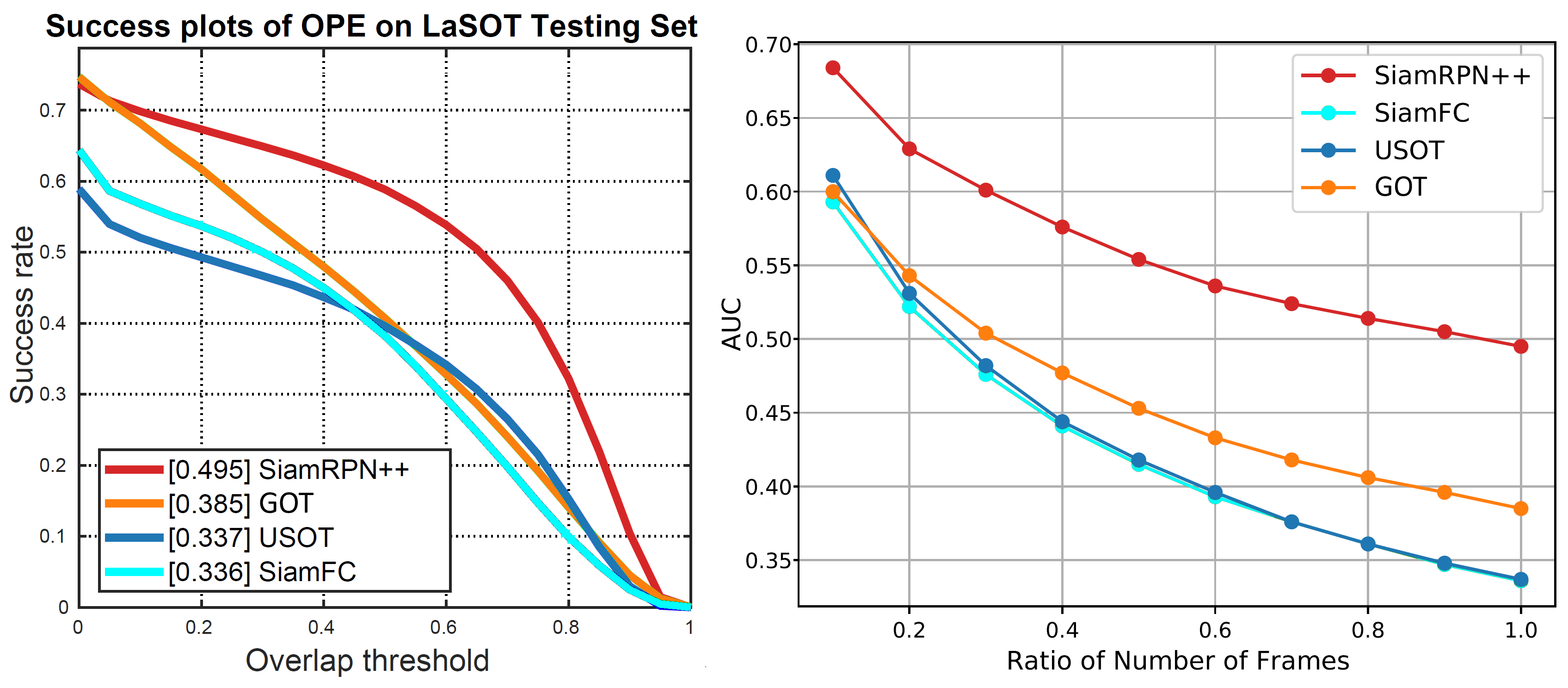}}
\caption{Performance comparison between GOT (orange), SiamRPN++ (red), SiamFC 
(cyan), and USOT (blue) on LaSOT in terms of the success rate plot (left) and
the AUC plot.} \label{fig:lasot_gap_plot}
\end{figure}

To analyze the limitations of GOT and gain a deeper understanding of the
contributions of supervision and offline pre-training, we compare the
performance of GOT and three DL trackers on the LaSOT dataset in
Fig.~\ref{fig:lasot_gap_plot}.  The three benchmarking trackers are
SiamRPN++ (a supervised DL tracker), USOT (an unsupervised DL tracker),
and SiamFC (a supervised DL tracker that does not have a regression
network as the previous two DL trackers).  The regression network is
offline pre-trained.  The left subfigure depicts the success rate as a
function of different overlap thresholds.  The right subfigure shows the
AUC values as a function of different video lengths with the full length
normalized to one. 

SiamRPN++ has the best performance among all. It is attributed to both
supervision and offline pre-training.  GOT ranks second in most
situations except for the following cases. GOT is slightly worse than
USOT when the overlap threshold is higher or at the beginning part of
videos.  It is conjectured that GOT can achieve decent shape estimation
but it may not be as effective as the offline pre-trained regression
network used by USOT in a tighter condition, i.e., a higher overlap
threshold or a shorter tracking memory. It is amazing to see that GOT
outperforms SiamFC across all thresholds and all video lengths. This
shows the importance of the regression network in DL trackers. 

\begin{figure}[htbp]
\centerline{\includegraphics[width=0.8\linewidth]{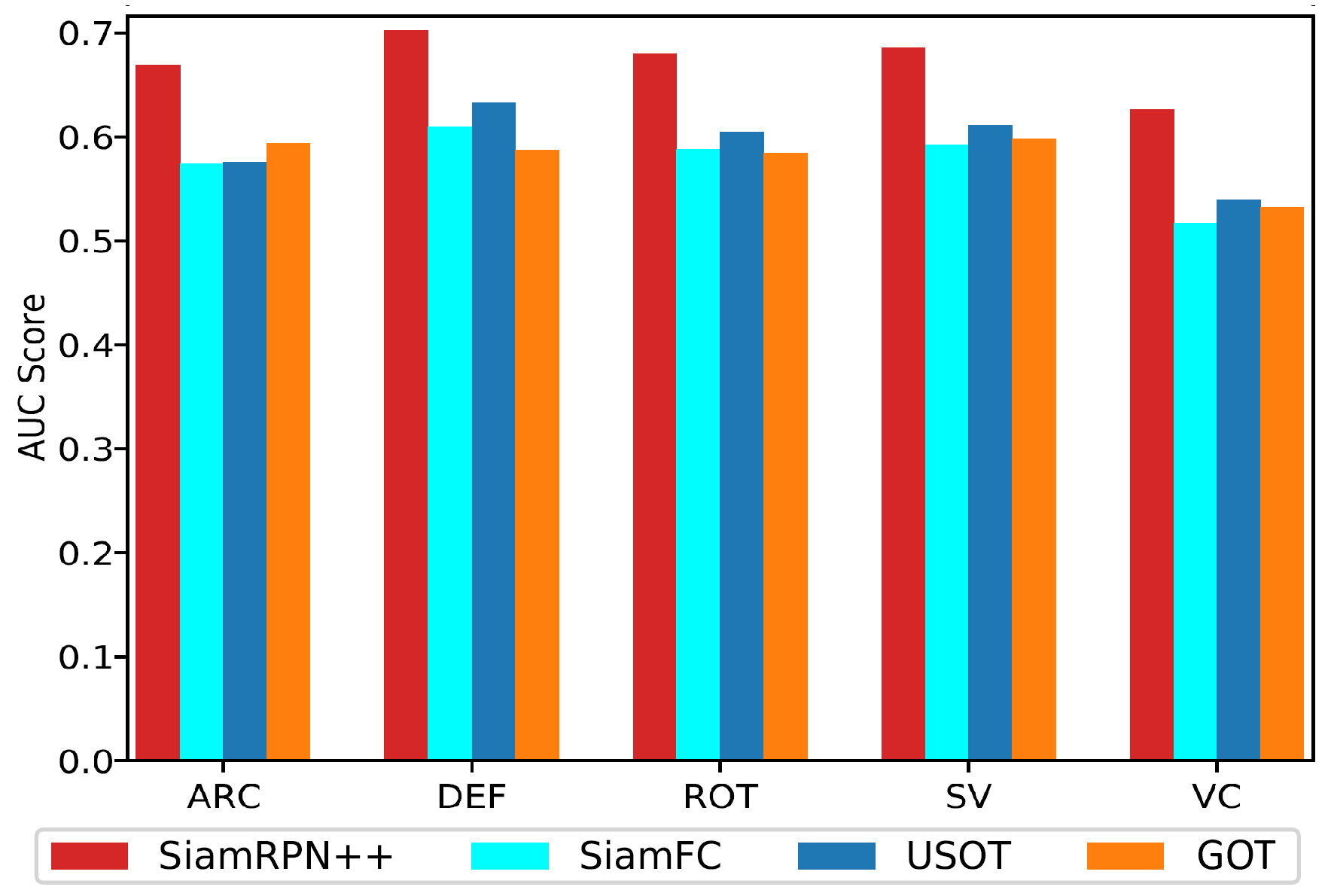}}
\caption{Deformation-related attribute study on LaSOT for the first
10\% frames in videos.} \label{fig:lasot_gap_att}
\end{figure}

To verify the above conjecture, we conduct the deformation-related
attribute study in a shorter tracking memory setting in
Fig.~\ref{fig:lasot_gap_att}, where only the first 10\% of all frames
(i.e., the ratio of frame numbers is 0.1) is examined.  GOT is better
than SiamFC in most attributes and worse than SiamRPN++ and USOT.  The
superiority of SiamFC over GOT in DEF and ROT indicates the power of
supervision in locating the object. 

\begin{figure}[htbp]
\centerline{\includegraphics[width=\linewidth]{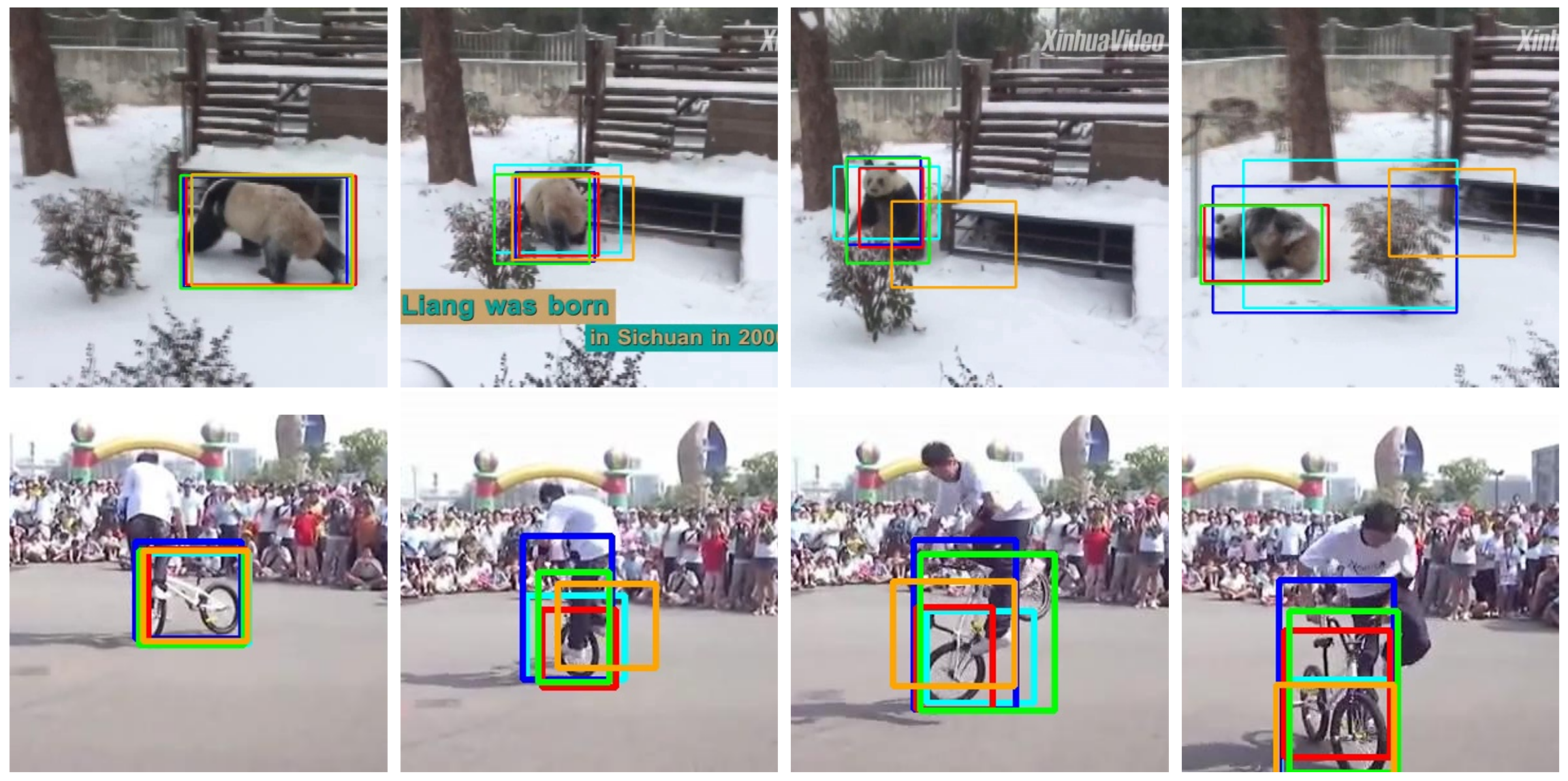}}
\caption{Two tracking examples from LaSOT: \textit{bear-4} (top) and
\textit{bicycle-2} (bottom), with boxes of the ground truth (green),
SiamRPN++ (red), SiamFC (cyan), USOT (blue), and GOT (orange), respectively.} 
\label{fig:lasot_gap_sample}
\end{figure}

We dive into two sequences where GOT's attributes are poorer and show
the tracking results in Fig.~\ref{fig:lasot_gap_sample}.  As shown in
these examples, GOT has difficulty in handling the following cases: (1)
the object has similar local features with the background, such as the
panda in the top sequence; and (2) the object under tracking does not
have a tight shape, such as the bike in the bottom sequence.  For the
first case, the local patch-based correlator in GOT can only capture
low-level visual similarities. It cannot distinguish the black color of
the panda and the background. For the second case, the bounding box
contains background patches inside, which become false positive
samples. Besides, the object has a few small individual parts whose
representations are not stable to reflect the appearance of the full
object. 

The motion may help if the object has obvious movement in the scene.
However, it may not help much if the object moves slowly or overlaps
with another object. These difficulties can be alleviated with
supervision or pre-training. That is, if a learning model is trained
with a rich set of object boxes, it can avoid such mistakes more easily.
As for the gap between supervised pre-training and unsupervised
pre-training, unsupervised methods use pseudo boxes generated randomly
or from the optical flow that usually contains noise. Thus, the offered
supervision is not as strong as ground truth labels. This explains why
USOT cannot distinguish between the panda and the bush and fails to
exclude the human body from the bike while SiamRPN++ does a good job. 

Supervised DL trackers usually do not distinguish different tracking
scenarios but tune a model to handle all cases to achieve high accuracy.
Their high computational complexity, large model sizes and heavy demand
on training data are costly.  In contrast, the proposed GOT system with
no offline pre-training can achieve decent tracking performance on general
videos. Possible ways to enhance the performance of lightweight trackers
include the design of better classifiers that have a higher level of
semantic meaning and more powerful regressors for better fusion of
predictions from various branches. 

\section{Conclusion and Future Work}\label{sec:conclusion}

A green object tracker (GOT) with a small model size, low inference
complexity and high tracking accuracy was proposed in this work.  GOT
contains a novel local patch-based correlator branch to enable more
flexible shape estimation and object re-identification. Furthermore, it
has a fusion tool that combines prediction outputs from the global
object-based correlator, the local patch-based correlator, and the
superpixel-based segmentator according to tracking dynamics over time.
Extensive experiments were conducted to compare the tracking performance
of GOT and state-of-the-art DL trackers. We hope that this work could
shed light on the role played by supervision and offline pre-training
and provide new insights into the design of effective and efficient
tracking systems.
 
Several future extensions can be considered. It is desired to develop
ways to identify different tracking scenarios since this information can
be leveraged to design a better tracking system. For example, it can
adopt tools of different complexity to strike a balance between model
complexity and tracking performance. Second, it can adopt different
fusion strategies to combine outputs from multiple decision branches for
more flexible and robust tracking performance. 

\appendices
\section{Model Size and Complexity Analysis of GOT}

The number of model parameters and the computational complexity analysis
of the proposed GOT system are analyzed.  For the latter, we compute the
floating point operations (flops) of the optimal implementation.
Whenever it is applicable, we offer the time complexity and provide a
rough estimation based on the running time on our local machine. The
complexity analysis is conducted for tracking a new frame in the
inference stage. Regardless of the original size of the frame, the
region of interest is always warped into a $L_B\times L_B = 60\times 60$
patch.  The major components of GOT include the global object-based
correlator (i.e., the DCF tracker), the local patch-based correlator
(i.e., the classification pipeline), the superpixel-based segmentator,
and the Markov Random Field (MRF) optimizer in the fuser.

\textbf{Global Object-based Correlator.} The DCF tracker involves
template matching via FFT and template updating via regression. The
template (feature map) dimension used in GOT is $(M,N,D)=(50,50,42)$.
The complexity of template updating is $\mathcal{O}(DMN \log{MN})$,
where $DMN \log{MN} \approx 1.19$M. The complexity of template matching
is at the same level. Furthermore, there is a background motion modeling
module in GUSOT to capture salient moving objects in the scene.  The
location of a certain point $(x_t,y_t)$ is estimated from its location
in frame $t-1$ via the following affine transformation,
\begin{eqnarray}
    x_t & = & a_0 x_{t-1} + b_0 y_{t-1} + c_0,\\
    y_t & = & a_1 x_{t-1} + b_1 y_{t-1} + c_1.
\end{eqnarray}
It is applied to every pixel in frame $I_{t-1}$ to get an estimation
$\hat{I}_t$ of frame $I_t$. Then, the motion residual map is calculated
as
\begin{equation}
    \Delta I = | \hat{I}_t - I_t |.
\end{equation}
The maximum dimension of $\Delta I$ is $(H,W)=(720,480)$ as images of a
larger size are downsampled. Flops for the affine transformation and the
residual map calculation are $8*H*W + H*W \approx 3.11$M. 

\textbf{Local Patch-based Correlator.} An input image of size $60\times
60$ is decomposed into overlapping blocks of size $8\times 8$ with a
stride equal to 2, which generates $((60-8)/2+1)^2=729$ blocks for
features extraction.  For Saab feature extraction, we apply the
one-layer Saab transform with filters of size $5 \times 5$ and stride
equal to 1. We keep the top 4 AC kernels from each of the PQR channels.
Thus, there are 3 DC color responses of size $3 \times 1\times 1$, and
12 AC responses of size $5 \times 5 \times 1$.  After feature
extraction, we apply the DFT feature selection method to reduce the
feature dimension to 50. Hence, the number of parameters is calculated
as $3\times3\textit{ (color kernels)} + 12\times25\textit{ (Saab
kernels)} + 50\textit{ (DFT feature selection index)} = 359.$
Since the Saab feature extraction process can be implemented as
3D convolutions as in neural networks, we follow the flops calculation
there to compute the model flops for the Saab transform. For a general 3D
convolution with $C_i$ input channels, $C_o$ filters of spatial size
$K_h\times K_w$ and output spatial size of $H_o\times W_o$, the flops is
calculated as
\begin{equation}
    F = (2\times C_i\times K_h\times K_w)\times H_o\times W_o\times C_o.
\end{equation}
If the filter is a mean filter, the complexity is further reduced as
\begin{align}
    F &= (C_i\times K_h\times K_w)\times H_o\times W_o\times C_o.
\end{align}
As given in Table \ref{tab:apdx}, the flops in computing the Saab
features with filter size $5\times 5$ at stride 1 for a block of size
$8\times 8\times 3$ is 11952. Then, the complexity for 729 blocks is
around 8.713M. We run this feature extraction process at most two times
at each frame. 

\begin{table}[htbp]
\caption{Flops of the Saab feature extraction for one spatial block
of size $8\times8$.}\label{tab:apdx}
\begin{center}
\renewcommand*{\arraystretch}{1.3}
\begin{tabular}{cccccccc}\hline
 Steps & $C_i$ & $K_h$ & $K_w$ & $H_o$ & $W_o$ & $C_o$ & Flops  \\ \hline
 Get mean color & 1 & 5 & 5 & 4 & 4 & 3 & 1200 \\
 RGB2PQR & 3 & 1 & 1 & 8 & 8 & 3 & 1152 \\
 Saab on P & 1 & 5 & 5 & 4 & 4 & 4 & 3200 \\
 Saab on Q & 1 & 5 & 5 & 4 & 4 & 4 & 3200 \\
 Saab on R & 1 & 5 & 5 & 4 & 4 & 4 & 3200 \\ \hline
 \textbf{Total} & & & & & & & \textbf{11952}\\
\hline
\end{tabular}
\end{center}
\end{table}

\begin{table}[htbp]
\caption{The model size and the computational complexity of the whole 
GOT system.}\label{tab:apdx_est_total}
\begin{center}
\renewcommand*{\arraystretch}{1.3}
\begin{tabular}{ccc}\hline
 Module & Num. of Params. & MFlops\\ \hline
 Global Correlator & 0 & 37.11\\
 Local Correlator & 2,199 & 18.12 \\
 Super-pixel segmentation & 0 & 1.13 \\ 
 MRF & 0 & 1.20 \\ \hline
 \textbf{Total} & \textbf{2,199} & \textbf{57.56} \\ \hline
\end{tabular}
\end{center}
\end{table}

\begin{table}[htbp]
\caption{The estimated flops for some special algorithmic modules.}\label{tab:apdx_est}
\begin{center}
\renewcommand*{\arraystretch}{1.3}
\begin{tabular}{ccc}\hline
 Algorithmic Modules & Complexity & MFlops  \\ \hline
 2D FFT\&IFFT & $\mathcal{O}(L_B^2 \log{L_B})$ & 0.072\\
 GMM & - & 1.634 \\
 DCF template related & $\mathcal{O}(DMN \log{MN})$ & 34 \\
 Super-pixel segmentation & $\mathcal{O}(L_B^2\log{L_B})$ & 1.132 \\ \hline
\end{tabular}
\end{center}
\end{table}

The XGBoost classifier has $N_{tree}=40$ trees with the maximum depth
$d_M=4$ (i.e., there are at most four tree levels excluding the root).
The maximum number of leaf nodes and parent nodes are $N_l=2^{d_M}$ and
$N_p=2^{d_M}-1$, respectively. Hence, the number of parameters is
bounded by $N_{tree}\times(2\times N_p + N_l) =40 \times (2\times15+16)
=1840$. The inference for 729 samples costs $d_M \times N_{tree} \times
729=4*40*729\approx0.117$M flops. The complexity of spatial alignment via 2D
FFT/IFFT is $\mathcal{O}(L_B^2 \log{L_B})$.  $L_B^2 \log_{2}{L_B}
\approx 0.021$M. The element-wise operation to get suppressed map takes
$L_B\times L_B=3600$ flops. The template update costs around $3\times
L_B\times L_B = 10800$ flops. 

\textbf{Felzenszwalb Superpixel Segmentator.} The complexity of the
superpixel segmentation algorithm is $\mathcal{O}(L_B^2\log{L_B})$, which roughly takes 1.13 MFlops. 

\textbf{MRF.} The adopted Markov Random Field optimizer has one
iteration only. Given an input image of size $(L_B,L_B,C)=(60,60,3)$, it
first learns the GMM models for foreground and background colors,
respectively, so that the foreground/background likelihood can be
calculated at each pixel. Then, around 20 element-wise matrix operations
are conducted to calculate the rough assignment of pixel labels. The
flops for matrix operations are $20*60*60=0.072M$. 

We summarize the model size (in the number of model parameters) and the
overall complexity (in flops) in Table \ref{tab:apdx_est_total}. Our
tracker has 2,199 model parameters and roughly 57.56 MFlops. It is
worthwhile to point out that the actual complexity of some special
modules such as FFT depend on the hardware implementation and
optimization. Some of them are given in Table \ref{tab:apdx_est}. 

\section*{Acknowledgment}

This material is based on research sponsored by US Army Research
Laboratory (ARL) under contract number W911NF2020157.  The U.S.
Government is authorized to reproduce and distribute reprints for
Governmental purposes notwithstanding any copyright notation thereon.
The views and conclusions contained herein are those of the authors and
should not be interpreted as necessarily representing the official
policies or endorsements, either expressed or implied, of US Army
Research Laboratory (ARL) or the U.S. Government. Computation for the
work was supported by the University of Southern California's Center for
Advanced Research Computing (CARC). 

\bibliographystyle{IEEEtran}
\bibliography{ref}

\begin{thebibliography}{10}
\providecommand{\url}[1]{#1}
\csname url@samestyle\endcsname
\providecommand{\newblock}{\relax}
\providecommand{\bibinfo}[2]{#2}
\providecommand{\BIBentrySTDinterwordspacing}{\spaceskip=0pt\relax}
\providecommand{\BIBentryALTinterwordstretchfactor}{4}
\providecommand{\BIBentryALTinterwordspacing}{\spaceskip=\fontdimen2\font plus
\BIBentryALTinterwordstretchfactor\fontdimen3\font minus
  \fontdimen4\font\relax}
\providecommand{\BIBforeignlanguage}[2]{{%
\expandafter\ifx\csname l@#1\endcsname\relax
\typeout{** WARNING: IEEEtran.bst: No hyphenation pattern has been}%
\typeout{** loaded for the language `#1'. Using the pattern for}%
\typeout{** the default language instead.}%
\else
\language=\csname l@#1\endcsname
\fi
#2}}
\providecommand{\BIBdecl}{\relax}
\BIBdecl

\bibitem{javed2022visual}
S.~Javed, M.~Danelljan, F.~S. Khan, M.~H. Khan, M.~Felsberg, and J.~Matas,
  ``Visual object tracking with discriminative filters and siamese networks: a
  survey and outlook,'' \emph{IEEE Transactions on Pattern Analysis and Machine
  Intelligence}, 2022.

\bibitem{lee2015road}
K.-H. Lee and J.-N. Hwang, ``On-road pedestrian tracking across multiple
  driving recorders,'' \emph{IEEE Transactions on Multimedia}, vol.~17, no.~9,
  pp. 1429--1438, 2015.

\bibitem{janai2020computer}
J.~Janai, F.~G{\"u}ney, A.~Behl, A.~Geiger \emph{et~al.}, ``Computer vision for
  autonomous vehicles: Problems, datasets and state of the art,''
  \emph{Foundations and Trends{\textregistered} in Computer Graphics and
  Vision}, vol.~12, no. 1--3, pp. 1--308, 2020.

\bibitem{xing2010multiple}
J.~Xing, H.~Ai, and S.~Lao, ``Multiple human tracking based on multi-view
  upper-body detection and discriminative learning,'' in \emph{2010 20th
  International Conference on Pattern Recognition}.\hskip 1em plus 0.5em minus
  0.4em\relax IEEE, 2010, pp. 1698--1701.

\bibitem{bolme2010visual}
D.~S. Bolme, J.~R. Beveridge, B.~A. Draper, and Y.~M. Lui, ``Visual object
  tracking using adaptive correlation filters,'' in \emph{2010 IEEE computer
  society conference on computer vision and pattern recognition}.\hskip 1em
  plus 0.5em minus 0.4em\relax IEEE, 2010, pp. 2544--2550.

\bibitem{danelljan2015learning}
M.~Danelljan, G.~Hager, F.~Shahbaz~Khan, and M.~Felsberg, ``Learning spatially
  regularized correlation filters for visual tracking,'' in \emph{Proceedings
  of the IEEE international conference on computer vision}, 2015, pp.
  4310--4318.

\bibitem{li2019siamrpn++}
B.~Li, W.~Wu, Q.~Wang, F.~Zhang, J.~Xing, and J.~Yan, ``Siamrpn++: Evolution of
  siamese visual tracking with very deep networks,'' in \emph{Proceedings of
  IEEE/CVF Computer Vision and Pattern Recognition}, 2019, pp. 4282--4291.

\bibitem{yan2021lighttrack}
B.~Yan, H.~Peng, K.~Wu, D.~Wang, J.~Fu, and H.~Lu, ``Lighttrack: Finding
  lightweight neural networks for object tracking via one-shot architecture
  search,'' in \emph{Proceedings of the IEEE/CVF Conference on Computer Vision
  and Pattern Recognition}, 2021, pp. 15\,180--15\,189.

\bibitem{shen2021distilled}
J.~Shen, Y.~Liu, X.~Dong, X.~Lu, F.~S. Khan, and S.~Hoi, ``Distilled siamese
  networks for visual tracking,'' \emph{IEEE Transactions on Pattern Analysis
  and Machine Intelligence}, vol.~44, no.~12, pp. 8896--8909, 2021.

\bibitem{blatter2023efficient}
P.~Blatter, M.~Kanakis, M.~Danelljan, and L.~Van~Gool, ``Efficient visual
  tracking with exemplar transformers,'' in \emph{Proceedings of the IEEE/CVF
  Winter Conference on Applications of Computer Vision}, 2023, pp. 1571--1581.

\bibitem{chen2022efficient}
X.~Chen, B.~Kang, D.~Wang, D.~Li, and H.~Lu, ``Efficient visual tracking via
  hierarchical cross-attention transformer,'' in \emph{European Conference on
  Computer Vision}.\hskip 1em plus 0.5em minus 0.4em\relax Springer, 2022, pp.
  461--477.

\bibitem{borsuk2022fear}
V.~Borsuk, R.~Vei, O.~Kupyn, T.~Martyniuk, I.~Krashenyi, and J.~Matas, ``Fear:
  Fast, efficient, accurate and robust visual tracker,'' in \emph{European
  Conference on Computer Vision}.\hskip 1em plus 0.5em minus 0.4em\relax
  Springer, 2022, pp. 644--663.

\bibitem{jung2022online}
I.~Jung, M.~Kim, E.~Park, and B.~Han, ``Online hybrid lightweight
  representations learning: Its application to visual tracking,'' \emph{arXiv
  preprint arXiv:2205.11179}, 2022.

\bibitem{aggarwal2023designing}
S.~Aggarwal, T.~Gupta, P.~K. Sahu, A.~Chavan, R.~Tiwari, D.~K. Prasad, and
  D.~K. Gupta, ``On designing light-weight object trackers through network
  pruning: Use cnns or transformers?'' in \emph{ICASSP 2023-2023 IEEE
  International Conference on Acoustics, Speech and Signal Processing
  (ICASSP)}.\hskip 1em plus 0.5em minus 0.4em\relax IEEE, 2023, pp. 1--5.

\bibitem{wang2021unsupervisedcvpr}
G.~Wang, Y.~Zhou, C.~Luo, W.~Xie, W.~Zeng, and Z.~Xiong, ``Unsupervised visual
  representation learning by tracking patches in video,'' in \emph{Proceedings
  of IEEE/CVF Computer Vision and Pattern Recognition}, 2021, pp. 2563--2572.

\bibitem{wu2021progressive}
Q.~Wu, J.~Wan, and A.~B. Chan, ``Progressive unsupervised learning for visual
  object tracking,'' in \emph{Proceedings of IEEE/CVF Computer Vision and
  Pattern Recognition}, 2021, pp. 2993--3002.

\bibitem{zheng2021learning}
J.~Zheng, C.~Ma, H.~Peng, and X.~Yang, ``Learning to track objects from
  unlabeled videos,'' in \emph{Proceedings of the IEEE/CVF International
  Conference on Computer Vision}, 2021, pp. 13\,546--13\,555.

\bibitem{shen2022unsupervised}
Q.~Shen, L.~Qiao, J.~Guo, P.~Li, X.~Li, B.~Li, W.~Feng, W.~Gan, W.~Wu, and
  W.~Ouyang, ``Unsupervised learning of accurate siamese tracking,'' in
  \emph{Proceedings of IEEE/CVF Computer Vision and Pattern Recognition}, 2022,
  pp. 8101--8110.

\bibitem{zhou2021uhp}
Z.~Zhou, H.~Fu, S.~You, C.~C. Borel-Donohue, and C.-C.~J. Kuo, ``{UHP-SOT}: An
  unsupervised high-performance single object tracker,'' in \emph{2021
  International Conference on Visual Communications and Image Processing
  (VCIP)}.\hskip 1em plus 0.5em minus 0.4em\relax IEEE, 2021, pp. 1--5.

\bibitem{zhou2022uhp}
Z.~Zhou, H.~Fu, S.~You, C.-C.~J. Kuo \emph{et~al.}, ``{UHP-SOT++}: An
  unsupervised lightweight single object tracker,'' \emph{APSIPA Transactions
  on Signal and Information Processing}, vol.~11, no.~1, 2022.

\bibitem{zhou2022gusot}
Z.~Zhou, H.~Fu, S.~You, and C.-C.~J. Kuo, ``Gusot: Green and unsupervised
  single object tracking for long video sequences,'' in \emph{2022 IEEE 24th
  International Workshop on Multimedia Signal Processing (MMSP)}.\hskip 1em
  plus 0.5em minus 0.4em\relax IEEE, 2022, pp. 1--6.

\bibitem{kuo2022green}
C.-C.~J. Kuo and A.~M. Madni, ``Green learning: Introduction, examples and
  outlook,'' \emph{Journal of Visual Communication and Image Representation},
  p. 103685, 2022.

\bibitem{henriques2014high}
J.~F. Henriques, R.~Caseiro, P.~Martins, and J.~Batista, ``High-speed tracking
  with kernelized correlation filters,'' \emph{IEEE transactions on pattern
  analysis and machine intelligence}, vol.~37, no.~3, pp. 583--596, 2014.

\bibitem{danelljan2015convolutional}
M.~Danelljan, G.~Hager, F.~Shahbaz~Khan, and M.~Felsberg, ``Convolutional
  features for correlation filter based visual tracking,'' in \emph{Proceedings
  of the IEEE international conference on computer vision workshops}, 2015, pp.
  58--66.

\bibitem{danelljan2016discriminative}
M.~Danelljan, G.~H{\"a}ger, F.~S. Khan, and M.~Felsberg, ``Discriminative scale
  space tracking,'' \emph{IEEE transactions on pattern analysis and machine
  intelligence}, vol.~39, no.~8, pp. 1561--1575, 2016.

\bibitem{danelljan2016beyond}
M.~Danelljan, A.~Robinson, F.~S. Khan, and M.~Felsberg, ``Beyond correlation
  filters: Learning continuous convolution operators for visual tracking,'' in
  \emph{European conference on computer vision}.\hskip 1em plus 0.5em minus
  0.4em\relax Springer, 2016, pp. 472--488.

\bibitem{bertinetto2016staple}
L.~Bertinetto, J.~Valmadre, S.~Golodetz, O.~Miksik, and P.~H. Torr, ``Staple:
  Complementary learners for real-time tracking,'' in \emph{Proceedings of the
  IEEE conference on computer vision and pattern recognition}, 2016, pp.
  1401--1409.

\bibitem{li2018learning}
F.~Li, C.~Tian, W.~Zuo, L.~Zhang, and M.-H. Yang, ``Learning spatial-temporal
  regularized correlation filters for visual tracking,'' in \emph{Proceedings
  of the IEEE conference on computer vision and pattern recognition}, 2018, pp.
  4904--4913.

\bibitem{xu2019learning}
T.~Xu, Z.-H. Feng, X.-J. Wu, and J.~Kittler, ``Learning adaptive discriminative
  correlation filters via temporal consistency preserving spatial feature
  selection for robust visual object tracking,'' \emph{IEEE Transactions on
  Image Processing}, vol.~28, no.~11, pp. 5596--5609, 2019.

\bibitem{li2020autotrack}
Y.~Li, C.~Fu, F.~Ding, Z.~Huang, and G.~Lu, ``Autotrack: Towards
  high-performance visual tracking for uav with automatic spatio-temporal
  regularization,'' in \emph{Proceedings of IEEE/CVF Computer Vision and
  Pattern Recognition}, 2020, pp. 11\,923--11\,932.

\bibitem{deng2009imagenet}
J.~Deng, W.~Dong, R.~Socher, L.-J. Li, K.~Li, and L.~Fei-Fei, ``Imagenet: A
  large-scale hierarchical image database,'' in \emph{2009 IEEE conference on
  computer vision and pattern recognition}.\hskip 1em plus 0.5em minus
  0.4em\relax Ieee, 2009, pp. 248--255.

\bibitem{lin2014microsoft}
T.-Y. Lin, M.~Maire, S.~Belongie, J.~Hays, P.~Perona, D.~Ramanan,
  P.~Doll{\'a}r, and C.~L. Zitnick, ``Microsoft coco: Common objects in
  context,'' in \emph{Computer Vision--ECCV 2014: 13th European Conference,
  Zurich, Switzerland, September 6-12, 2014, Proceedings, Part V 13}.\hskip 1em
  plus 0.5em minus 0.4em\relax Springer, 2014, pp. 740--755.

\bibitem{russakovsky2015imagenet}
O.~Russakovsky, J.~Deng, H.~Su, J.~Krause, S.~Satheesh, S.~Ma, Z.~Huang,
  A.~Karpathy, A.~Khosla, M.~Bernstein \emph{et~al.}, ``Imagenet large scale
  visual recognition challenge,'' \emph{International journal of computer
  vision}, vol. 115, pp. 211--252, 2015.

\bibitem{real2017youtube}
E.~Real, J.~Shlens, S.~Mazzocchi, X.~Pan, and V.~Vanhoucke,
  ``Youtube-boundingboxes: A large high-precision human-annotated data set for
  object detection in video,'' in \emph{proceedings of the IEEE Conference on
  Computer Vision and Pattern Recognition}, 2017, pp. 5296--5305.

\bibitem{muller2018trackingnet}
M.~Muller, A.~Bibi, S.~Giancola, S.~Alsubaihi, and B.~Ghanem, ``Trackingnet: A
  large-scale dataset and benchmark for object tracking in the wild,'' in
  \emph{Proceedings of the European conference on computer vision (ECCV)},
  2018, pp. 300--317.

\bibitem{huang2019got}
L.~Huang, X.~Zhao, and K.~Huang, ``Got-10k: A large high-diversity benchmark
  for generic object tracking in the wild,'' \emph{IEEE transactions on pattern
  analysis and machine intelligence}, vol.~43, no.~5, pp. 1562--1577, 2019.

\bibitem{ma2015hierarchical}
C.~Ma, J.-B. Huang, X.~Yang, and M.-H. Yang, ``Hierarchical convolutional
  features for visual tracking,'' in \emph{Proceedings of the IEEE
  international conference on computer vision}, 2015, pp. 3074--3082.

\bibitem{qi2016hedged}
Y.~Qi, S.~Zhang, L.~Qin, H.~Yao, Q.~Huang, J.~Lim, and M.-H. Yang, ``Hedged
  deep tracking,'' in \emph{Proceedings of the IEEE conference on computer
  vision and pattern recognition}, 2016, pp. 4303--4311.

\bibitem{danelljan2017eco}
M.~Danelljan, G.~Bhat, F.~Shahbaz~Khan, and M.~Felsberg, ``Eco: Efficient
  convolution operators for tracking,'' in \emph{Proceedings of the IEEE
  conference on computer vision and pattern recognition}, 2017, pp. 6638--6646.

\bibitem{bertinetto2016fully}
L.~Bertinetto, J.~Valmadre, J.~F. Henriques, A.~Vedaldi, and P.~H. Torr,
  ``Fully-convolutional siamese networks for object tracking,'' in
  \emph{European conference on computer vision}.\hskip 1em plus 0.5em minus
  0.4em\relax Springer, 2016, pp. 850--865.

\bibitem{li2018high}
B.~Li, J.~Yan, W.~Wu, Z.~Zhu, and X.~Hu, ``High performance visual tracking
  with siamese region proposal network,'' in \emph{Proceedings of the IEEE
  conference on computer vision and pattern recognition}, 2018, pp. 8971--8980.

\bibitem{tao2016siamese}
R.~Tao, E.~Gavves, and A.~W. Smeulders, ``Siamese instance search for
  tracking,'' in \emph{Proceedings of the IEEE conference on computer vision
  and pattern recognition}, 2016, pp. 1420--1429.

\bibitem{zhu2018distractor}
Z.~Zhu, Q.~Wang, B.~Li, W.~Wu, J.~Yan, and W.~Hu, ``Distractor-aware siamese
  networks for visual object tracking,'' in \emph{Proceedings of the European
  Conference on Computer Vision (ECCV)}, 2018, pp. 101--117.

\bibitem{zhang2019deeper}
Z.~Zhang and H.~Peng, ``Deeper and wider siamese networks for real-time visual
  tracking,'' in \emph{Proceedings of the IEEE/CVF conference on computer
  vision and pattern recognition}, 2019, pp. 4591--4600.

\bibitem{wang2021transformer}
N.~Wang, W.~Zhou, J.~Wang, and H.~Li, ``Transformer meets tracker: Exploiting
  temporal context for robust visual tracking,'' in \emph{Proceedings of
  IEEE/CVF Computer Vision and Pattern Recognition}, 2021, pp. 1571--1580.

\bibitem{chen2021transformer}
X.~Chen, B.~Yan, J.~Zhu, D.~Wang, X.~Yang, and H.~Lu, ``Transformer tracking,''
  in \emph{Proceedings of IEEE/CVF Computer Vision and Pattern Recognition},
  2021, pp. 8126--8135.

\bibitem{felzenszwalb2004efficient}
P.~F. Felzenszwalb and D.~P. Huttenlocher, ``Efficient graph-based image
  segmentation,'' \emph{International journal of computer vision}, vol.~59,
  no.~2, pp. 167--181, 2004.

\bibitem{chen2020pixelhop++}
Y.~Chen, M.~Rouhsedaghat, S.~You, R.~Rao, and C.-C.~J. Kuo, ``Pixelhop++: A
  small successive-subspace-learning-based (ssl-based) model for image
  classification,'' in \emph{2020 IEEE International Conference on Image
  Processing (ICIP)}.\hskip 1em plus 0.5em minus 0.4em\relax IEEE, 2020, pp.
  3294--3298.

\bibitem{yang2022supervised}
Y.~Yang, W.~Wang, H.~Fu, C.-C.~J. Kuo \emph{et~al.}, ``On supervised feature
  selection from high dimensional feature spaces,'' \emph{APSIPA Transactions
  on Signal and Information Processing}, vol.~11, no.~1, 2022.

\bibitem{chen2016xgboost}
T.~Chen and C.~Guestrin, ``Xgboost: A scalable tree boosting system,'' in
  \emph{Proceedings of the 22nd acm sigkdd international conference on
  knowledge discovery and data mining}, 2016, pp. 785--794.

\bibitem{wang2021unsupervised}
N.~Wang, W.~Zhou, Y.~Song, C.~Ma, W.~Liu, and H.~Li, ``Unsupervised deep
  representation learning for real-time tracking,'' \emph{International Journal
  of Computer Vision}, vol. 129, no.~2, pp. 400--418, 2021.

\bibitem{7001050}
Y.~Wu, J.~Lim, and M.-H. Yang, ``Object tracking benchmark,'' \emph{IEEE
  Transactions on Pattern Analysis and Machine Intelligence}, vol.~37, no.~9,
  pp. 1834--1848, 2015.

\bibitem{hadfield2016visual}
S.~Hadfield, R.~Bowden, and K.~Lebeda, ``The visual object tracking vot2016
  challenge results,'' \emph{Lecture Notes in Computer Science}, vol. 9914, pp.
  777--823, 2016.

\bibitem{fan2019lasot}
H.~Fan, L.~Lin, F.~Yang, P.~Chu, G.~Deng, S.~Yu, H.~Bai, Y.~Xu, C.~Liao, and
  H.~Ling, ``Lasot: A high-quality benchmark for large-scale single object
  tracking,'' in \emph{Proceedings of IEEE/CVF Computer Vision and Pattern
  Recognition}, 2019, pp. 5374--5383.

\bibitem{valmadre2018long}
J.~Valmadre, L.~Bertinetto, J.~F. Henriques, R.~Tao, A.~Vedaldi, A.~W.
  Smeulders, P.~H. Torr, and E.~Gavves, ``Long-term tracking in the wild: A
  benchmark,'' in \emph{Proceedings of the European conference on computer
  vision (ECCV)}, 2018, pp. 670--685.

\bibitem{dai2020high}
K.~Dai, Y.~Zhang, D.~Wang, J.~Li, H.~Lu, and X.~Yang, ``High-performance
  long-term tracking with meta-updater,'' in \emph{Proceedings of the IEEE/CVF
  conference on computer vision and pattern recognition}, 2020, pp. 6298--6307.

\bibitem{kalal2011tracking}
Z.~Kalal, K.~Mikolajczyk, and J.~Matas, ``Tracking-learning-detection,''
  \emph{IEEE transactions on pattern analysis and machine intelligence},
  vol.~34, no.~7, pp. 1409--1422, 2011.

\bibitem{lukevzivc2019fucolot}
A.~Luke{\v{z}}i{\v{c}}, L.~{\v{C}}. Zajc, T.~Voj{\'\i}{\v{r}}, J.~Matas, and
  M.~Kristan, ``Fucolot--a fully-correlational long-term tracker,'' in
  \emph{Computer Vision--ACCV 2018: 14th Asian Conference on Computer Vision,
  Perth, Australia, December 2--6, 2018, Revised Selected Papers, Part II
  14}.\hskip 1em plus 0.5em minus 0.4em\relax Springer, 2019, pp. 595--611.

\end{thebibliography}

\end{document}